\documentclass{article}

\usepackage{microtype}
\usepackage{graphicx}
\usepackage{subfigure}
\usepackage{booktabs} %

\usepackage{multirow}
\usepackage{tabularx}
\usepackage{adjustbox}
\usepackage{makecell}
\usepackage{amssymb}%
\usepackage{pifont}%
\newcommand{\cmark}{\ding{51}}%
\newcommand{\xmark}{\ding{55}}%
\usepackage{booktabs}  %
\usepackage{dsfont} %
\usepackage{amsmath} %
\usepackage{booktabs}
\usepackage{float}
\usepackage{lipsum}  %
\usepackage{enumitem}
\usepackage{xcolor}
\usepackage{amsthm}

\newtheorem{theorem}{Theorem}%
\newtheorem{proposition}[theorem]{Proposition}
\newcommand{\variance}[1]{{\color[HTML]{AF1E1E} #1}}
\newcommand{\bias}[1]{{\color[HTML]{00D59D} #1}}
\theoremstyle{plain} %

\usepackage{amsmath,amsfonts,bm}

\def\eqref#1{equation~\ref{#1}}

\def\1{\bm{1}}

\def\vzero{{\bm{0}}}
\def\vone{{\bm{1}}}

\def\vtheta{{\bm{\theta}}}
\def\va{{\bm{a}}}

\def\vv{{\bm{v}}}
\def\vw{{\bm{w}}}
\def\vx{{\bm{x}}}
\def\vy{{\bm{y}}}

\def\mA{{\bm{A}}}

\def\mH{{\bm{H}}}
\def\mI{{\bm{I}}}

\def\mQ{{\bm{Q}}}

\def\mX{{\bm{X}}}

\DeclareMathAlphabet{\mathsfit}{\encodingdefault}{\sfdefault}{m}{sl}
\SetMathAlphabet{\mathsfit}{bold}{\encodingdefault}{\sfdefault}{bx}{n}

\def\gD{{\mathcal{D}}}

\def\gN{{\mathcal{N}}}
\def\gO{{\mathcal{O}}}

\def\gR{{\mathcal{R}}}

\newcommand{\E}{\mathbb{E}}

\newcommand{\R}{\mathbb{R}}

\def\mDelta{{\bm{\Delta}}}
\def\mPi{{\bm{\Pi}}}
\def\rvvarepsilon{{\mathbf{\varepsilon}}}
\def\rvgamma{{\bm{\gamma}}}
\def\rvdelta{{\bm{\delta}}}

\usepackage{hyperref}

\usepackage[accepted]{icml2024}

\usepackage{amsmath}
\usepackage{amssymb}
\usepackage{mathtools}
\usepackage{amsthm}

\usepackage[capitalize,noabbrev]{cleveref}

\theoremstyle{plain}

\theoremstyle{definition}

\theoremstyle{remark}

\usepackage[textsize=tiny]{todonotes}
\usepackage[T1]{fontenc}

\icmltitlerunning{Pi-DUAL: Using privileged information to distinguish clean from noisy labels}

\begin{document}

\twocolumn[
\icmltitle{Pi-DUAL: Using privileged information to distinguish clean from noisy labels}

\icmlsetsymbol{equal}{*}

\begin{icmlauthorlist}
\icmlauthor{Ke Wang}{epfl}
\icmlauthor{Guillermo Ortiz-Jimenez}{deepmind,atepfl}
\icmlauthor{Rodolphe Jenatton}{Bioptimus,atdeepmind}
\icmlauthor{Mark Collier}{google}
\icmlauthor{Efi Kokiopoulou}{google}
\icmlauthor{Pascal Frossard}{epfl}
\end{icmlauthorlist}

\icmlaffiliation{epfl}{École Polytechnique Fédérale de Lausanne (EPFL)}
\icmlaffiliation{atepfl}{Work done while at EPFL}
\icmlaffiliation{deepmind}{Google DeepMind}
\icmlaffiliation{atdeepmind}{Work done while at Google DeepMind}
\icmlaffiliation{google}{Google Research}
\icmlaffiliation{Bioptimus}{Bioptimus}

\icmlcorrespondingauthor{Ke Wang}{k.wang@epfl.ch}

\icmlkeywords{Machine Learning, Label Noise, Privileged Information, ICML}

\vskip 0.3in
]

\printAffiliationsAndNotice{} %

\begin{abstract}
Label noise is a pervasive problem in deep learning that often compromises the generalization performance of trained models. Recently, leveraging privileged information (PI) -- information available only during training but not at test time -- has emerged as an effective approach to mitigate this issue. Yet, existing PI-based methods have failed to consistently outperform their no-PI counterparts in terms of preventing overfitting to label noise. To address this deficiency, we introduce Pi-DUAL, an architecture designed to harness PI to distinguish clean from wrong labels. Pi-DUAL decomposes the output logits into a prediction term, based on conventional input features, and a noise-fitting term influenced solely by PI. A gating mechanism steered by PI adaptively shifts focus between these terms, allowing the model to implicitly separate the learning paths of clean and wrong labels. Empirically, Pi-DUAL achieves significant performance improvements on key PI benchmarks (e.g., $+6.8\%$ on ImageNet-PI), establishing a new state-of-the-art test set accuracy. Additionally, Pi-DUAL is a potent method for identifying noisy samples post-training, outperforming other strong methods at this task.  Overall, Pi-DUAL is a simple, scalable and practical approach for mitigating the effects of label noise in a variety of real-world scenarios with PI. \looseness=-1
\end{abstract}

\section{Introduction}
\label{submission}

Many deep learning models are trained on large noisy datasets, 
as obtaining cleanly labeled datasets at scale
can be expensive and time consuming~\citep{snow-etal-2008-cheap,sheng_2008}. However, the presence of label noise in the training set tends to damage generalization performance as it forces the model to learn spurious associations between the input features and the noisy labels \citep{ZhangBengioHardtRechtVinyals.17, arpit2017closer}. To mitigate the negative effects of label noise,
recent methods have primarily tried to prevent overfitting to the noisy labels, often utilising the observation that neural networks tend to first learn the clean labels before memorizing the wrong ones~\citep{maennel_2020,baldock_2021}.
For instance, these methods
include filtering out incorrect labels, correcting them, or enforcing regularization on the training dynamics~\citep{han2018co,liu2020early,li2020dividemix}. Other works, instead, try to
capture the noise structure in an input-dependent fashion~\citep{patrini2017making,liu2022robust,collier2022transfer,collier2023hetxl}.
\looseness=-1

The above methods are however designed for a standard supervised learning setting, where models are tasked to learn
an association
between input features $\vx \in \R^d$ and targets $y\in\{1,\dots,K\}$ (assuming $K$ classes) from a training set of pairs $\{(\vx_i,\tilde{y}_i)\}_{i\in[n]}$ of features and 
(possibly)
noisy labels $\tilde{y}\in\{1,\dots K\}$. 
As a result, they need to model the noise in the targets as a function of $\vx$.
Yet, in many practical situations, the mistakes introduced during the annotation process may not solely depend on the input $\vx$, but rather be mostly explained by annotation-specific side information, such as the experience of the annotator or the attention they paid while annotating.
For this reason, a recent line of work~\citep{lupi_vapnik,collier2022transfer,ortiz2023does} has proposed to use \emph{privileged information} (PI) to mitigate the affects of label noise.
PI is defined as additional features available at training time but not at test time. It can include annotation features such as the annotator ID, the amount of time to provide the label, or their experience.

Remarkably, having access to PI at training time, even when it is not available at test time, has been shown to be an effective tool for dealing with instance-dependent label noise. Most notably, \citet{ortiz2023does} showed that by exploiting PI it is possible to activate positive learning shortcuts to memorize, and therefore explain away, noisy training samples, thereby improving generalization. Nevertheless, and perhaps surprisingly, current PI-based methods do not systematically  outperform no-PI baselines in the presence of label noise, making them a less competitive alternative in certain cases~\citep{ortiz2023does}.\looseness=-1

\begin{figure*}[t]
\centering
\includegraphics[width=\textwidth]{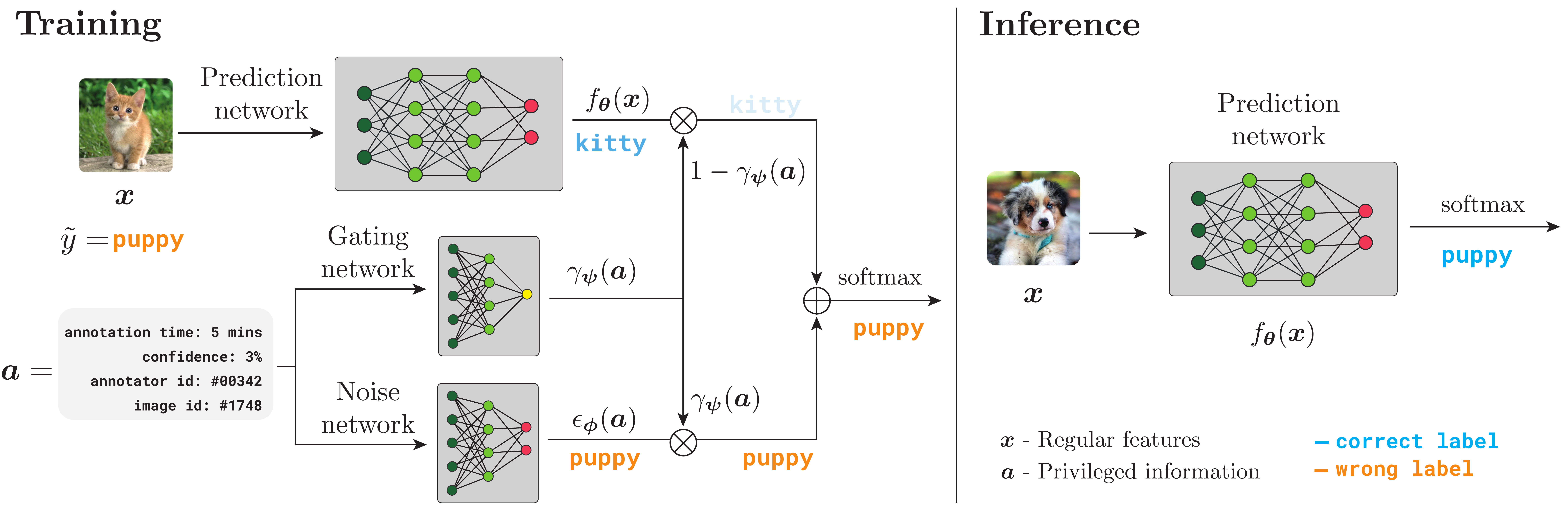}
\caption{\textbf{Illustration of the architecture of Pi-DUAL.} (Left) During training, Pi-DUAL fits the noisy target label $\tilde{y}$ combining the output of a prediction network (which takes the regular features $\vx$ as input) and a noise network (which takes the PI $\va$ as input). The outputs of these sub-networks are weighted based on the output of a gating network (which also has $\va$ as input) and then passed through a $\operatorname{softmax}$ operator to obtain the predictions. (Right) During inference, when only $\vx$ is available, Pi-DUAL does not need access to PI and simply uses the prediction network to predict the clean target $y$.\looseness=-1}

\label{fig:introductory_fig} 
\end{figure*}

In this work, we aim to improve the performance of PI strategies
by proposing a new PI-guided noisy label architecture: \textbf{Pi-DUAL}, a \textbf{P}rivileged \textbf{I}nformation network to \textbf{D}istinguish \textbf{U}ntrustworthy \textbf{A}nnotations and \textbf{L}abels. Specifically, during training, we propose to decompose the output logits into a weighted combination of a prediction term, that depends only on the regular features $\vx$, and a noise-fitting term, that depends only on the PI features $\va \in \R^p$. Pi-DUAL toggles between these terms using a gating mechanism, also solely a function of $\va$, that decides if a sample should be learned primarily by the prediction network, or explained away by the noise network (see Fig.~\ref{fig:introductory_fig}). This dual sub-network design adaptively routes the clean and wrong labels through the prediction and noise networks so that they are fit based on $\vx$ or $\va$, respectively.
This protects the prediction network from overfitting to the label noise. Pi-DUAL is simple to implement, effective, and can be trained end-to-end with minimal modifications to existing training pipelines. Unlike some previous methods Pi-DUAL also scales to training on very large datasets. Finally, in public benchmarks for learning with label noise, Pi-DUAL achieves state-of-the-art results on datasets with rich PI features ($+4.5\%$ on CIFAR-10H, $+1.3\%$ on ImageNet-PI (low-noise) and $+6.8\%$ on ImageNet-PI (high-noise)); and performs on par with previous methods on benchmarks with weak PI or no PI at all, despite not being specifically designed to work in these regimes.

Overall, the main contributions of our work are:
\begin{itemize}%

    \item We present Pi-DUAL, a novel PI method to combat label noise based on a dual path architecture that implicitly separates the noisy fitting path from the clean prediction path during training.\looseness=-1

    \item We show that Pi-DUAL achieves strong performance on noisy label benchmarks without sacrificing scalability, outperforming previous state-of-the-art methods when given access to high-quality PI features.

    \item We provide a simple yet effective method to detect wrong labels in the training set using the prediction network of Pi-DUAL, achieving strong detection performance.
    
\end{itemize}

In summary, our work advances the state-of-the-art in noisy label learning by effectively leveraging privileged information through the novel Pi-DUAL architecture. Pi-DUAL can be easily integrated into any learning pipeline, requires minimal hyperparameters, and can be trained end-to-end in a single stage. Overall, Pi-DUAL is a scalable and practical approach for mitigating the effects of label noise in a variety of real-world scenarios with PI.

\section{Related Work}
\label{seg:related_work}

\begin{table*}[ht]
\caption{Comparison of different representative methods to learn with label noise \emph{vs} Pi-DUAL on several design axes: ability to leverage PI, ability to explicitly model the noise signal, parameter scalability, and whether training requires multiple models and training stages. Scalability indicates whether the number of parameters for the method remains constant regardless of the number of samples or the number of classes in the training set.\looseness=-1}
\label{table:baselines}
\begin{center}
\adjustbox{max width=\textwidth}{%
\begin{tabular}{@{}ccccccc@{}}
\toprule
& & & & & \\[-1em]
  Methods & Leverage PI &  \begin{tabular}[c]{@{}c@{}}Explicit\\ noise modeling \end{tabular}  & \begin{tabular}[c]{@{}c@{}}Scalability w.r.t.\\ num. of samples \end{tabular}  & \begin{tabular}[c]{@{}c@{}}Scalability w.r.t.\\ num. of classes \end{tabular}  & \begin{tabular}[c]{@{}c@{}}Training\\ complexity \end{tabular}  \\ \midrule
& & & & & \\[-1em] %
                       Forward-T \citep{patrini2017making} & \xmark & \cmark & \cmark & \xmark & 1 model, 2 stages &  \\ 
                       Co-Teaching \citep{han2018co} & \xmark & \xmark & \cmark & \cmark & 2 models, 1 stage &  \\ 
                       Divide-Mix \citep{li2020dividemix} & \xmark & \xmark & \cmark & \cmark & 2 models, 1 stage &  \\
                       ELR \citep{liu2020early} & \xmark & \xmark & \xmark & \xmark & 1 model, 1 stage &  \\ 
                       SOP \citep{liu2022robust} & \xmark & \cmark & \xmark & \xmark & 1 model, 1 stage & \\
                       HET-XL \citep{collier2023hetxl} & \xmark & \cmark & \cmark & \cmark & 1 model, 1 stage &  \\ 
                       Distill. PI \citep{lopez2015unifying} & \cmark & \xmark & \cmark & \cmark & 2 model, 2 stage & \\
                       AFM \citep{collier2022transfer} & \cmark & \xmark & \cmark & \cmark & 1 model, 1 stage & \\
                       TRAM++ \citep{ortiz2023does} & \cmark & \xmark & \cmark & \cmark & 1 model, 1 stage & \\ \midrule
                       Pi-DUAL (Ours) & \cmark & \cmark & \cmark & \cmark & 1 model, 1 stage & \\ 
                       \bottomrule
\end{tabular}%
}
\end{center}
\end{table*}

Noisy label methods mostly fall into two broad categories: 
those that explicitly model the noise signal, and those that rely on implicit network dynamics to correct or ignore the wrong labels~\citep{song2022learning}.  Noise modeling techniques aim to learn the function that governs the noisy annotation process explicitly during training, inverting it during inference to obtain the clean labels. Some methods model the annotation function using a transition matrix~\citep{patrini2017making}; others model uncertainty via a heteroscedastic noise term~\citep{het,collier2023hetxl}; and recently, some works explicitly parameterize the label error signal as a vector for each sample in the training set~\citep{tanaka2018joint,yi2019probabilistic,liu2022robust}. Implicit-dynamics based approaches, on the other hand, operate under the assumption that wrong labels are harder to learn than the correct labels~\citep{ZhangBengioHardtRechtVinyals.17,maennel_2020}. Using this intuition, different methods have come up with different heuristics to correct~\citep{jiang2018mentornet,han2018co,yu2019does} or downweight~\citep{liu2020early,menon2020can,bai2021understanding} the influence of wrong labels during training. This has sometimes led to very complex methods that require multiple stages of training~\citep{patrini2017making,bai2021understanding,albert2023your,wang2022promix}, higher computational cost~\citep{han2018co,jiang2018mentornet,han2018co,yu2019does}, and many additional parameters that do not scale well to large datasets~\citep{yi2019probabilistic,liu2020early,liu2022robust}.\looseness=-1

The introduction of privileged information (PI) offers an alternative dimension to tackle the noisy label problem~\citep{hernandez2014mind,lopez2015unifying,collier2022transfer}. In this regard,
\citet{ortiz2023does} showed that most PI methods work as implicit-dynamics approaches. They rely on the use of PI to enable learning shortcuts, to avoid memorizing the incorrect labels using the regular features.
Moreover, these approaches are attractive for their scalability, as they usually avoid the introduction of extra training stages or parameters. However, current PI methods can sometimes lag behind in performance with respect to no-PI baselines. The main reason is that these methods still try to learn the noise predictive distribution $p(\tilde{y}|\vx)$ by marginalizing $\va$ in $p(\tilde{y}|\vx,\va)$, when they should actually aim to learn the clean distribution $p(y|\vx)$ directly. However, prior PI methods do not have an explicit mechanism to identify or correct
the wrong labels.\looseness=-1

Our proposed method, Pi-DUAL, tries to circumvent these issues by explicitly modeling the \textit{clean distribution}, exploiting the ability of PI to distinguish clean and wrong labels. Our design allows Pi-DUAL to scale effectively across large datasets and diverse class distributions, while maintaining high performance and low training complexity as seen in Tab.~\ref{table:baselines}. We further note that our design is reminiscent of mixtures of experts (MoE) that were shown to be a competitive architecture for language modeling~\citep{shazeer2017outrageously} and computer vision~\citep{riquelme2021scaling}. By analogy, we can see Pi-DUAL as an MoE containing a single MoE layer with two heterogeneous experts---the prediction and noise networks---located at the logits of the model and with a dense gating.%

\section{Pi-DUAL}

\subsection{Noise Modeling}
\label{sec:background}

In traditional supervised learning, we typically assume that there exists a groundtruth function $f^\star:\mathcal{X}\to\mathcal{Y}$ which maps input features $\bm x\in\mathcal{X}$ to labels $y\in\mathcal{Y}$ where $\mathcal{X}=\mathbb{R}^d$ and $\mathcal{Y}=\{1,\dots,K\}$. However, the labels in real-world scenarios are usually gathered via a noisy annotation process.

In this work, we model this annotation process as a function of some, possibly unknown, side information $\bm a\in\mathcal{A}$, which explains away the noise from the training labels. This side information could be anything, from the experience of the annotator, to their intrinsic motivation. The important modeling aspect is that given this side information one should be able to tell whether a label is incorrect or not, and the type of mistake that was made. We can model this process mathematically as a function $h:\mathcal{X}\times\mathcal{A}\to\mathcal{Y}$ that maps the input features and the side information to the noisy human label $\tilde{y}$. We assume that the mistakes in the annotation process depend only on $\va$, i.e.,
\begin{equation}
\tilde{y} = h(\vx,\va)= [1-\gamma(\bm a)]f^\star(\vx) + \gamma(\bm a) \epsilon(\va)
\label{eq:generative_model}
\end{equation}
Here $\gamma:\mathcal{A}\to\{0,1\}$ acts as a switch between clean and wrong labels, and $\epsilon:\mathcal{A}\to\mathcal{Y}$ models the incorrect labelling function. Consequently, the training dataset $\mathcal{D}$ consists of two types of training samples $\mathcal{D}_{\text{correct}} = \{(\vx, \tilde{y})\in\mathcal{D}\;|\; \tilde{y} = f^\star(\vx)\}$ and $\mathcal{D}_{\text{wrong}} = \{(\vx, \tilde{y})\in\mathcal{D}\;|\;\tilde{y}= \epsilon(\va)\}$. In this regard, when training a network to map $\vx$ to $\tilde{y}$ on $\mathcal{D} = \mathcal{D}_{\text{correct}} \cup \mathcal{D}_{\text{wrong}}$, we are effectively asking it to learn two different target functions, where only one of them depends on $\vx$, which forces the network to memorize part of the training data and hurts its generalization~\citep{ZhangBengioHardtRechtVinyals.17}.\looseness=-1

In practice, however, we will not have access to the exact side information, and we will be able to rely at most on meta-data, and PI, about the annotation process. That is, we consider a learning problem in which our training data consists of triplets $(\vx,\tilde{y};\va)$ where $\va\in\R^p$ is a vector of PI features such as high latency features related to the annotation process, e.g., annotator experience, or even a randomly assigned unique vector introduced to model unobserved features~\citep{ortiz2023does}. We present here our method that uses this setting to explicitly model $h$ and learn effectively in the presence of large amounts of label noise in the training set.

\subsection{Method Description}
\label{sec:method}

Based on the noise model, we propose Pi-DUAL, a novel PI-based architecture designed to mimic the generative noise model proposed in Eq. (\ref{eq:generative_model}). Specifically, during training, Pi-DUAL factorizes its output logits into two terms, i.e.,
\begin{equation}
h_{\bm\theta,\bm\phi,\bm\psi}(\vx,\va)=[1-\gamma_{\bm\psi}(\va)]f_{\bm\theta}(\vx) + \gamma_{\bm\psi}(\va)\epsilon_{\bm\phi}(\va),
\label{eq:model}
\end{equation}
where $f_{\bm\theta}:\mathcal{X}\to\R^c$ represents a \textit{prediction network} tasked with approximating the ground truth labelling function $f^\star$ and $\epsilon_{\bm\phi}:\mathcal{A}\to\R^c$ a \textit{noise network}, modeling the noise signal $\epsilon$. Here, $\gamma_{\bm\phi}$ denotes a \textit{gating network} tasked with learning the switching mechanism $\gamma$, where we apply a sigmoid activation function to the output to restrict $\gamma_{\bm\psi}(\bm a)$ to be within in $[0,1]$. Moreover, following the recommendations of \citet{ortiz2023does}, we augment the available PI features with a unique random identifier for each training sample to help the network explain away the missing factors of the noise using this identifier. The dimension of this vector, known as random PI length, is the only additional hyperparameter we tune for Pi-DUAL. During inference, when PI is not available, Pi-DUAL relies solely on $f_{\bm\theta}(\vx)$ to predict the clean label $y$ (see Fig. \ref{fig:introductory_fig}). 

The dual gated logit structure of Pi-DUAL is reminiscent of sparsely-gated mixture of experts which also factorize its predictions at the logit level, albeit providing the same input $\vx$ to each expert~\citep{shazeer2017outrageously}. Pi-DUAL instead provides $\vx$ and $\va$ to different networks, which effectively decouples learning the task-specific samples and the noise-specific samples with different features. Indeed, assuming that the incorrect labels are independent of $\vx$ and that the noise is only a function of the PI $\va$, there will always be a natural tendency by the network to use $\epsilon_{\bm\phi}(\va)$ to explain away those labels that it cannot easily learn with $f_{\bm\theta}(\vx)$. The gating network $\gamma_{\bm\phi}$ facilitates this separation by utilizing the discriminative power of the PI to guide this process. In Sec.~\ref{sec:ablation}, we ablate all these elements of the architecture to show that they all contribute to learning the clean labels.\looseness=-1

Pi-DUAL has multiple advantages over previous PI methods like TRAM or AFM~\citep{collier2022transfer}. Indeed, previous methods tend to directly expose the no-PI term $f_{\bm\theta}(\vx)$ to the noisy labels, e.g., through $\mathcal{L}(f_{\bm\theta}(\vx),\tilde{y})$, which can thus lead to an overfitting to the noisy labels based on $\vx$. In contrast, Pi-DUAL instead solves
\begin{equation}
\setlength{\abovedisplayskip}{8pt}
\setlength{\belowdisplayskip}{8pt}
   \underset{\bm\theta,\bm\phi,\bm\psi}{\operatorname{min}}\quad \sum_{ (\vx,\tilde{y};\bm a)\in\mathcal{D}}\mathcal{L}\left( \operatorname{softmax}\left(h_{\bm\theta,\bm\phi,\bm\psi}(\vx,\va)\right), \tilde{y}\right),
\end{equation}
and never explicitly forces $f_{\bm\theta}(\vx)$ to fit all $\tilde{y}$'s (we validate the loss design in Appendix~\ref{app:pidual_tram_loss}). Our design allows the model to predict clean label for all training samples without incurring loss penalty, as it can fit the residual noise signal with $\epsilon_{\bm\phi}(\bm a)$.
In Sec.~\ref{sec:dynamics} we analyze in detail these dynamics.\looseness=-1

Another important advantage of Pi-DUAL is that it explicitly learns to model noise signal in training set. This makes it more interpretable than implicit-dynamics methods like TRAM, and puts it on par with state-of-the-art noise-modeling methods. However, as Pi-DUAL can leverage PI to model noise signal, it exhibits a much better noise detection performance than no-PI methods, while at the same time allowing it to scale to datasets with millions of datapoints, as it does not require to store individual parameters for each sample in the training set to effectively learn the label noise.\looseness=-1

\subsection{Noise Detection}
\label{sec:noise_detection}

Finally, we provide a simple noise detection method based on Pi-DUAL: After training, we collect confidence estimates of the prediction network on the observed noisy labels $\tilde{y}$, i.e., with
$\operatorname{softmax}(f_{\theta}(\vx))[\tilde{y}]$,
for the training samples and threshold the confidences to distinguish the correctly and incorrectly labeled examples. Indeed, we show that because the prediction network $f_{\bm\theta}$ of Pi-DUAL only learns to confidently predict clean labels $y$ during training without having to memorize the wrong labels, its confidence on the noisy labels is a very good proxy for a noise indicator: If its confidence on an observed label is high, then it is highly likely that the sample is correctly labeled, i.e., $\tilde{y}=y$; but if it is low, then probably the label $\tilde{y}$ is wrong. In Sec.~\ref{sec:detection}, we compare Pi-DUAL to other state-of-the-art methods to detect wrong labels.

\subsection{Theoretical Insights}

To further support the design of Pi-DUAL described in Eq. (\ref{eq:model}), we study the theoretical behavior of the predictor $h_{\bm\theta,\bm\phi,\bm\psi}(\vx,\va)$ within a simplified linear regression setting. More specifically, we consider the setting where the clean and noisy targets are respectively generated from two Gaussian distributions $\gN(\vx^\top \vw^\star, \sigma^2)$ and $\gN(\va^\top \vv^\star, \sigma^2)$, for two weight vectors $(\vw^\star, \vv^\star)$ parameterizing linearly their means. In this tractable setting, we show that Pi-DUAL is a robust estimator in the presence of label noise as its risk depends less severely on the number of wrong labels.

We compare two estimators, Pi-DUAL and an ordinary least squares estimator (OLS) that ignores the side information $\va$. We summarize below the main insights of our analysis, and provide the details in Appendix \ref{app:risk_analysis}.

\begin{theorem}[Informal]
Consider $n$ samples from the above Gaussian models with targets $\vy = \rvgamma^\star \mX \vw^\star + (\mI - \rvgamma^\star) \mA \vv^\star + \rvvarepsilon$. The contributions of the standard and PI features are respectively $\mX \vw^\star \in \R^n$ and $\mA \vv^\star \in \R^n$, 
while $\rvgamma^\star \in \{0, 1\}^{n \times n}$ is a diagonal mask that indicates which contribution each entry in $\vy$ corresponds to.
Denoting $\rvdelta^\star= \mX \vw^\star - \mA \vv^\star \in \R^n$, it can be shown that the risk of the OLS estimator has a bias term scaling with $\gO((\mI - \rvgamma^*)\rvdelta^\star)$, while the risk of Pi-DUAL using an arbitrary diagonal mask $\rvgamma \in \{0, 1\}^{n \times n}$ has a bias term that depends on $\gO((\rvgamma^\star - \rvgamma) \rvdelta^\star)$, which only scales with the number of disagreements with respect to the ground-truth $\rvgamma^\star$. \looseness=-1
\end{theorem}

We show with this theorem that in terms of their abilities to generalize on \textit{clean targets}---as measured by their risks~\citep{bach2021learning}---Pi-DUAL exhibits a more robust behavior. In particular, while the risk of OLS tends to be proportional to the number of wrong labels $|\gD_\text{wrong}|$, Pi-DUAL has a risk that more gracefully scales with respect to the number of examples that the gates $\gamma_{\bm\psi}$ fail to identify. Our experiments in Sec.~\ref{sec:gates} show that, in practice, the gates learned by Pi-DUAL typically manage to identify the clean and wrong labels.\looseness=-1

\begin{table*}[ht]
\caption{Test accuracy of different methods on noisy label datasets with PI. We report mean and standard deviation accuracy over multiple runs with the best hyperparameters and early-stopping.}
\label{table:results}
\begin{center}
    
\begin{tabular}{@{}ccccccc@{}}
\toprule
& & & & & \\[-1em] 
\multicolumn{2}{c}{Methods}  & \begin{tabular}[c]{@{}c@{}}CIFAR-10H\\ (worst) \end{tabular} & \begin{tabular}[c]{@{}c@{}}CIFAR-10N\\ (worst)\end{tabular} & \begin{tabular}[c]{@{}c@{}}CIFAR-100N \\ (fine)\end{tabular} & \begin{tabular}[c]{@{}c@{}}ImageNet-PI\\ (low-noise)\end{tabular} & \begin{tabular}[c]{@{}c@{}}ImageNet-PI\\ (high-noise)\end{tabular} \\ \midrule
& & & & & \\[-1em] 
\multirow{4}{*}{\rotatebox[origin=c]{90}{No-PI}} 
                       &  Cross-entropy & $51.1_{\color{gray}\pm2.2}$ & $80.6_{\color{gray}\pm0.2}$ & $60.4_{\color{gray}\pm0.5}$ & $68.2_{\color{gray}\pm0.2}$ & $47.2_{\color{gray}\pm0.2}$ \\
                       &  ELR  & $48.5_{\color{gray}\pm1.4}$ & $\textbf{86.6}_{\color{gray}\pm0.7}$ & $\textbf{64.0}_{\color{gray}\pm0.3}$ & - & - \\ 
                       &  HET  & $50.8_{\color{gray}\pm1.4}$ & $81.9_{\color{gray}\pm0.4}$ & $60.8_{\color{gray}\pm0.4}$ & $69.4_{\color{gray}\pm0.1}$ & $51.9_{\color{gray}\pm0.0}$ \\ 
                       &  SOP  & $51.3_{\color{gray}\pm1.9}$  & $85.0_{\color{gray}\pm0.8}$  & $61.9_{\color{gray}\pm0.6}$  & - &  - \\ 
                       \midrule
\multirow{3}{*}{\rotatebox[origin=c]{90}{PI}} & TRAM & $64.9_{\color{gray}\pm0.8}$ & $80.5_{\color{gray}\pm0.5}$ & $59.7_{\color{gray}\pm0.3}$ & $69.4_{\color{gray}\pm0.2}$ & $54.0_{\color{gray}\pm0.1}$ \\ 
                       &  TRAM++ & $66.8_{\color{gray}\pm0.3}$ & $83.9_{\color{gray}\pm0.2}$ & $61.1_{\color{gray}\pm0.2}$ & $69.5_{\color{gray}\pm0.0}$ & $53.8_{\color{gray}\pm0.3}$ \\ 
                       &  AFM  & $64.0_{\color{gray}\pm0.6}$ & $82.0_{\color{gray}\pm0.3}$ & $60.0_{\color{gray}\pm0.2}$ & $70.3_{\color{gray}\pm0.0}$ & $55.3_{\color{gray}\pm0.2}$ \\ \midrule
& & & & & \\[-1em]
\multicolumn{2}{c}{Pi-DUAL (Ours)} & $\textbf{71.3}_{\color{gray}\pm3.3}$ & $84.9_{\color{gray}\pm0.4}$ & $\textbf{64.2}_{\color{gray}\pm0.3}$ & $\textbf{71.6}_{\color{gray}\pm0.1}$ & $\textbf{62.1}_{\color{gray}\pm0.1}$ \\ \bottomrule
\end{tabular}
\end{center}
\end{table*}

\section{Experimental Results}\label{sec:experiments}

We now validate the effectiveness of Pi-DUAL on several public noisy label benchmarks with PI and compare it extensively to other algorithms. We show that Pi-DUAL achieves (a) state-of-the-art results on clean test accuracy and noise detection tasks (especially when there is good PI available) and (b) scales up to datasets with millions of examples.\looseness=-1

\subsection{Experimental Settings}

Our experimental settings follow the benchmarking practices laid out by \citet{ortiz2023does}. In particular, we use the same architectures, training schedules and  public codebase~\citep{nado2021uncertainty} to perform all our experiments. In terms of baseline choices, in order to achieve a fair comparison, we 
compare Pi-DUAL to our own implementations of the methods in Tab.~\ref{table:baselines} that use only one model and one stage of training\footnote{We do not run ELR and SOP on ImageNet-PI as they require 1 billion extra parameters (see Appendix~\ref{app:baselines}).\looseness=-1}. Moreover, to further ensure fairness,
we use on each dataset the same architecture and the same training strategy across all compared methods.
For each result, we perform a grid search over hyperparameters. Notably, while other methods require tuning at least two additional hyperparameters on top of the cross-entropy baseline; Pi-DUAL only requires tuning the random PI length, making its tuning budget
much smaller. We use a \textit{noisy validation set}, held-out from the training set, to select the best hyperparameters and report results over the clean test set. We provide more details on our hyper-parameter tuning strategy and other experimental settings in Appendix~\ref{sec:experimental_details}, and a computational cost analysis in Appendix~\ref{app:computational_cost}.\looseness=-1 

\begin{figure*}[ht]
\centering
\includegraphics[width=\textwidth]{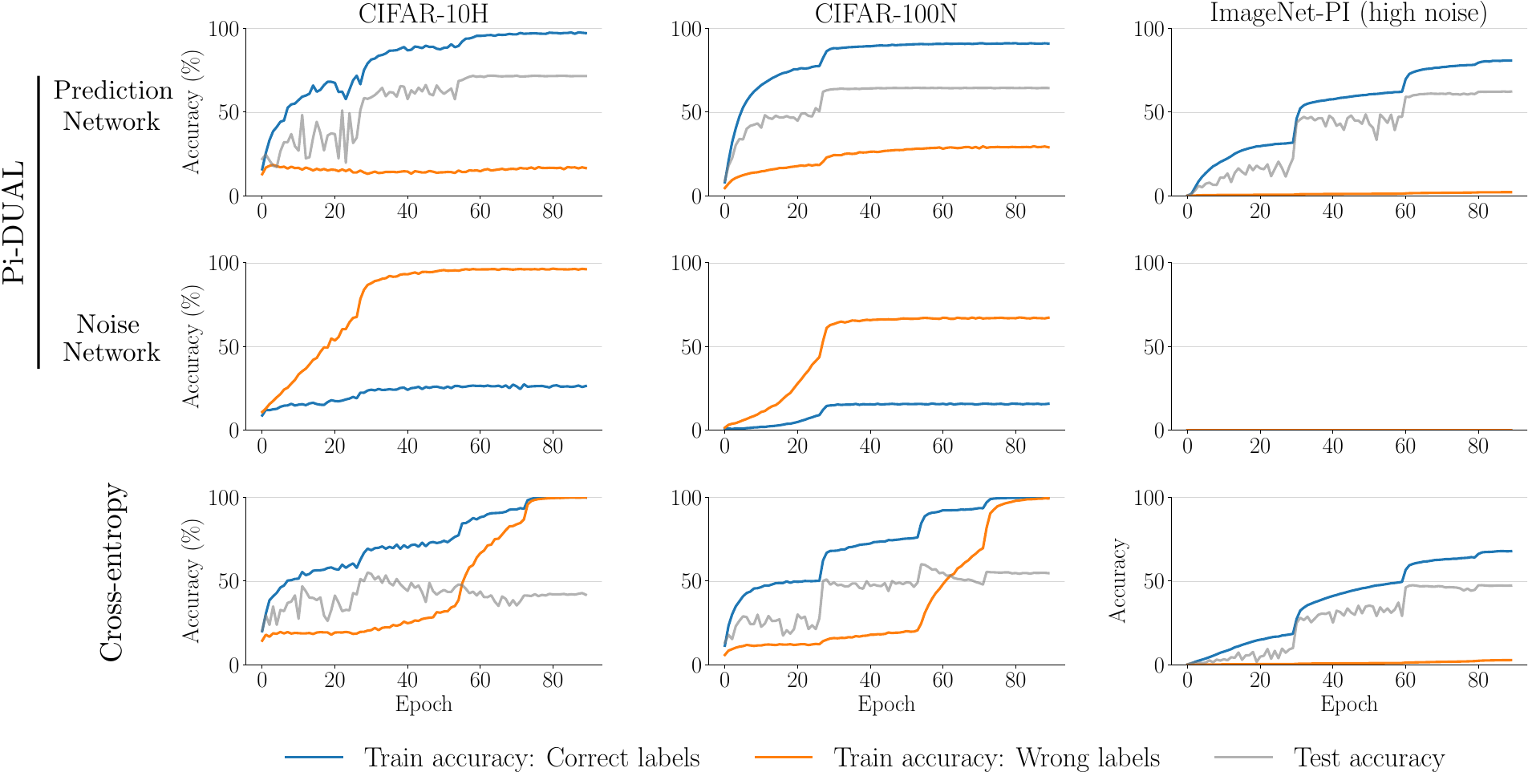}
\caption{Training curves of Pi-DUAL and cross-entropy baseline on different datasets. The first two rows show the training dynamics of prediction network and noise network respectively.We plot separately the training accuracy on clean and wrong labels and test accuracy\protect\footnotemark.}
\vspace{0.8em}
\label{fig:train_curve} 
\end{figure*}

Pi-DUAL does not require to use early stopping to achieve strong results as it does not suffer from overfitting issues (see Fig.~\ref{fig:train_curve}). However, early stopping is essential to achieve good performance for the other methods. Hence, we always report results at the epoch with the best accuracy on the noisy validation set. In Appendix~\ref{app:no_early_stopping}, we provide results for all methods without early stopping.\looseness=-1

Our experiments are conducted on five noisy datasets with realistic label noise, either derived from a noisy human annotation process or produced by imperfect model predictions. A summary of the main features of each datasets is shown in Appendix~\ref{app:dataset}.
Importantly, we note
that CIFAR-10H~\citep{cifarh}, ImageNet-PI (low-noise) and ImageNet-PI (high-noise) all have excellent-quality PI features at the sample level that seem to capture important information of the annotation process. On the other hand, CIFAR-10N and CIFAR-100N~\citep{cifarn} provide aggregated PI, in the form of averages over batches of samples, which may not have enough resolution to distinguish clean and wrong labels at the sample level~\citep{ortiz2023does}. Despite this, Pi-DUAL still performs comparatively to no-PI methods on those datasets.\looseness=-1

We further provide in Appendix \ref{app:pidual_plus} the results for Pi-DUAL+, a stronger version of Pi-DUAL boosted with advanced regularization techniques. It is competitive against state-of-art semi-supervised learning based methods, such as Divide-Mix \cite{li2020dividemix} and SOP+ \citep{liu2022robust}.

\subsection{Predicting Clean Labels} 

Tab.~\ref{table:results} reports the test accuracy of Pi-DUAL compared to previous noisy label methods, averaged over 5 and 3 random seeds for CIFAR and ImageNet-PI, respectively. As we can see, Pi-DUAL achieves state-of-the-art performance on the three datasets with high quality PI. It improves by $4.5\%$ over the most competitive PI baseline on CIFAR-10H and by $20$ points over the best performing no-PI methods. It also achieves a $1.3$ point and $6.8$ point lead on ImageNet-PI low-noise and high-noise, respectively. These are remarkable results given the 1000 classes in ImageNet-PI and the scale of these datasets. Indeed, they show that Pi-DUAL can effectively leverage PI in these settings to distinguish between correct and wrong labels during training,
while learning the clean labels with the prediction network.\looseness=-1

On the other hand, on the two datasets with low quality PI, we observe that Pi-DUAL achieves better results than previous PI methods by more than $3$ points on CIFAR-100N. It also performs comparatively with no-PI methods, even though the quality of the PI does not allow to properly distinguish between clean and wrong labels (see Sec.~\ref{sec:gates}).

\begin{table*}[t]
\caption{AUC of different noise detection methods based on confidence thresholding of the network predictions on noisy labels or thresholding of the gating network's output (for Pi-DUAL).}
\label{table:auc_number}
\begin{center}
\adjustbox{max width=0.8\textwidth}{%
\begin{tabular}{@{}cccccc@{}}
\toprule
& & & & & \\[-1em] %
  Methods & \begin{tabular}[c]{@{}c@{}}CIFAR-10H\\ (worst) \end{tabular} & \begin{tabular}[c]{@{}c@{}}CIFAR-10N\\ (worst)\end{tabular} & \begin{tabular}[c]{@{}c@{}}CIFAR-100N \\ (fine)\end{tabular} & \begin{tabular}[c]{@{}c@{}}ImageNet-PI\\ (low-noise)\end{tabular} & \begin{tabular}[c]{@{}c@{}}ImageNet-PI\\ (high-noise)\end{tabular} \\ \midrule
& & & & & \\[-1em] %
                      Cross-entropy & $0.810$ & $0.951$ & $0.883$ & $0.935$ & $0.941$ \\ 
                      ELR & $0.745$ & $\textbf{0.968}$ & $0.876$ & - & - \\ 
                      SOP & $0.808$ & $0.964$ & $0.889$ & - & - \\ 
                      TRAM++ & $0.834$ & $0.955$ & $0.883$ & $0.937$ & $0.959$ \\\midrule
& & & & & \\[-1em] %
                      Pi-DUAL (conf.) & $0.954$ & $0.962$ & $\textbf{0.911}$ & $\textbf{0.953}$ & $\textbf{0.986}$ \\
                      Pi-DUAL (gate) & $\textbf{0.982}$ & $0.808$ & $0.726$ & $0.952$ & $\textbf{0.986}$ \\ 
\bottomrule
\end{tabular}%
}
\end{center}
\end{table*}

\begin{figure*}[ht]
\centering
\includegraphics[width=0.9\textwidth]{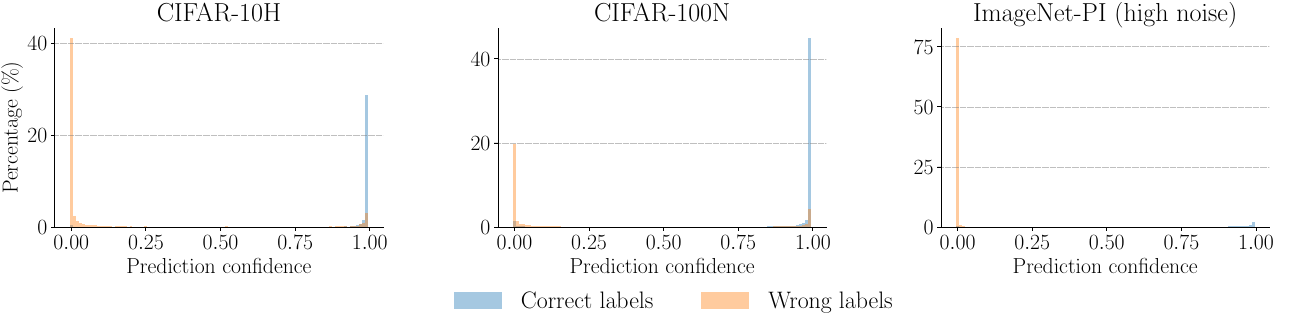}
\caption{Distribution for the prediction network's confidence on the observed noisy labels for several datasets, separated by correctly and wrongly labeled samples.}
\label{fig:confidence_distribution_main} 
\end{figure*}

\begin{figure*}[ht]
\centering
\includegraphics[width=0.9\textwidth]{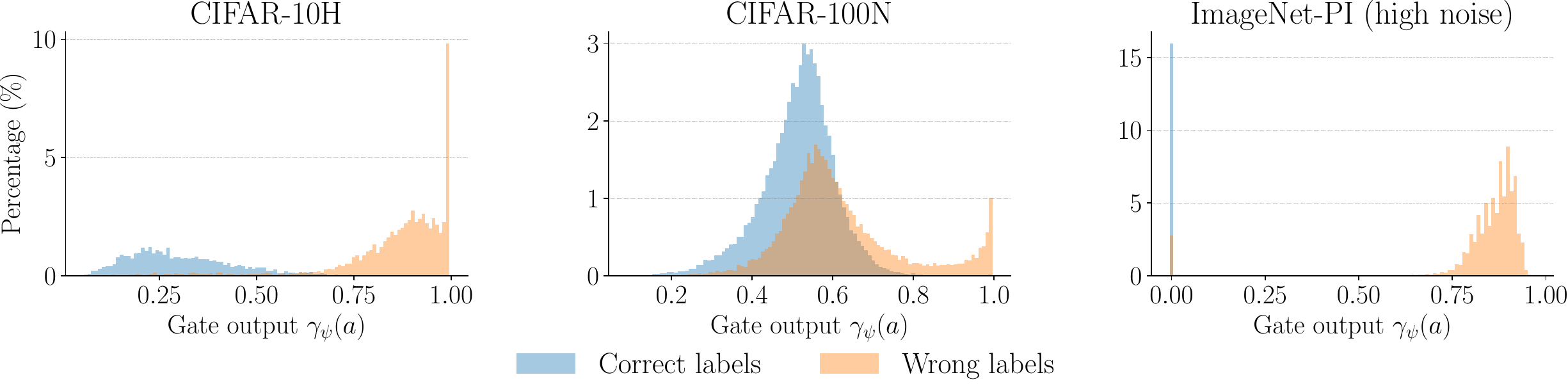}
\caption{Distributions of $\gamma_{\bm\psi}(\va)$ over training samples with correct and wrong labels on several datasets.\looseness=-1}
\label{fig:gate_distribution} 
\end{figure*}

\subsection{Detection of Wrong Labels} \label{sec:detection}

We validate the ability of Pi-DUAL to detect the wrong labels in the training set, allowing practitioners to relabel those instances, or filter them out in future runs. 

Tab.~\ref{table:auc_number} shows the area under the receiver operating characteristic curve (AUC)
obtained by applying our confidence-based noise detection method described in Section~\ref{sec:noise_detection}, compared with different methods on all PI benchmarks. As we can see, Pi-DUAL achieves the best results by a large margin in all datasets except CIFAR-10N (where it performs comparatively to the best method). These performance gains are a clear sign that Pi-DUAL can effectively minimize the amount of overfitting of the prediction network to the noisy labels. In most cases, the prediction network has a very low confidence (near 0\%) on the wrong labels, while having a very high confidence (near 100\%) on the correct labels. 
We show the distribution of the prediction confidences in Fig.~\ref{fig:confidence_distribution_main} for CIFAR-10H, CIFAR-100N, and ImageNet-PI (high noise) and two other datasets in Appendix~\ref{app:confidence_distribution}, where we observe that the prediction confidence is clearly separated over samples with correct and wrong labels.%

In our experiments, we observe that the simple confidence thresholding is a strong detection method across all datasets. Meanwhile, we also evaluated the ability of the gating network $\gamma_\psi$ in detecting the wrong labels. As shown in Tab.~\ref{table:auc_number}, thresholding the gate's outputs is also an effective method for noise detection, which can even outperform confidence thresholding on certain datasets, i.e., CIFAR-10H. However, we observe that the performance of gate thresholding suffers more than confidence thresholding on datasets with low-resolution PI. As we will see in Sec.~\ref{sec:gates}, this is due to the fact that, in those datasets, the gating network cannot exploit the PI to discriminate easily between correct and wrong labels. Still, this does not prevent the prediction network from learning the clean distribution, and thus its detection ability does not suffer as much. Choosing which of the two methods to use is, in general, a dataset-dependent decision: If there is good PI available gate thresholding achieves the best results, but confidence thresholding performs well overall, so we recommend it as a default choice.\looseness=-1

\begin{figure*}[ht]
\centering
\includegraphics[width =1.05\textwidth]{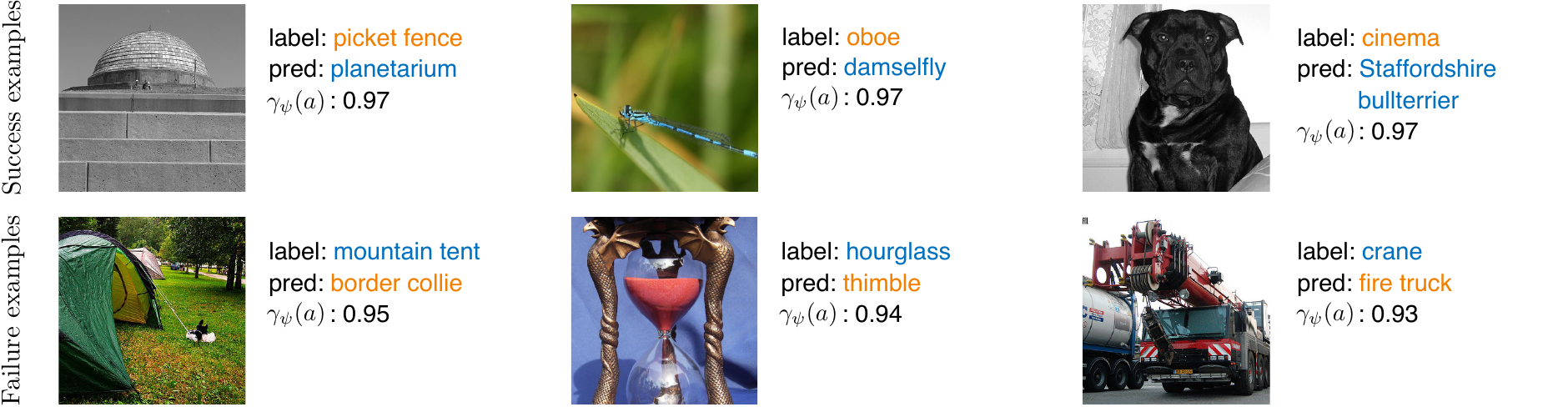}
\caption{Examples of ImageNet-PI images that the gating network suggests are mislabeled. The first row shows samples with actually wrongly annotated labels, and the second row shows examples with correct labels but assumed to be wrong by the gating network. Here, ``label" denotes the annotation label $\tilde{y}$ and ``pred" the prediction by $f_{\bm{\theta}}$.\looseness=-1}
\label{fig:visual_example} 
\end{figure*}

\begin{table*}[htbp]
\caption{Test accuracy of various ablation studies over Pi-DUAL on the different PI datasets.}
\label{table:ablation_structure}
\begin{center}
\adjustbox{max width=\textwidth}{%
\begin{tabular}{@{}cccccc@{}}
\toprule
& & & & & \\[-1em] %
  Ablations & \begin{tabular}[c]{@{}c@{}}CIFAR-10H\\ (worst) \end{tabular} & \begin{tabular}[c]{@{}c@{}}CIFAR-10N\\ (worst)\end{tabular} & \begin{tabular}[c]{@{}c@{}}CIFAR-100N \\ (fine)\end{tabular} & \begin{tabular}[c]{@{}c@{}}ImageNet-PI\\ (low-noise)\end{tabular} & \begin{tabular}[c]{@{}c@{}}ImageNet-PI\\ (high-noise)\end{tabular} \\ \midrule
& & & & & \\[-1em] %
                      Cross-entropy & $51.1_{\color{gray}\pm2.2}$ & $80.6_{\color{gray}\pm0.2}$ & $60.4_{\color{gray}\pm0.5}$ & $68.2_{\color{gray}\pm0.2}$ & $47.2_{\color{gray}\pm0.2}$ \\
                       Pi-DUAL & $\textbf{71.3}_{\color{gray}\pm3.3}$ & $\textbf{84.9}_{\color{gray}\pm0.4}$ & $\textbf{64.2}_{\color{gray}\pm0.3}$ & $\textbf{71.6}_{\color{gray}\pm0.1}$ & $\textbf{62.1}_{\color{gray}\pm0.1}$\\ \midrule
& & & & & \\[-1em] %
                        (no gating network) & $61.5_{\color{gray}\pm1.2}$ & $\textbf{84.5}_{\color{gray}\pm0.2}$ & $59.0_{\color{gray}\pm0.2}$ & $67.9_{\color{gray}\pm0.1}$ & $47.8_{\color{gray}\pm0.8}$ \\
                        (no noise network) & $59.7_{\color{gray}\pm3.6}$ & $82.4_{\color{gray}\pm1.0}$ & $59.7_{\color{gray}\pm0.3}$ & $\textbf{71.6}_{\color{gray}\pm0.2}$ & $\textbf{62.3}_{\color{gray}\pm0.1}$ \\ 
                       (gate in prob. space) & $62.2_{\color{gray}\pm1.3}$ & $81.6_{\color{gray}\pm0.8}$ & $59.4_{\color{gray}\pm1.1}$ & $71.0_{\color{gray}\pm0.1}$ & $60.4_{\color{gray}\pm0.1}$ \\
                       (only random PI) & $53.5_{\color{gray}\pm2.2}$ & $83.7_{\color{gray}\pm1.3}$ & $61.8_{\color{gray}\pm0.3}$ & $68.4_{\color{gray}\pm0.1}$ & $47.0_{\color{gray}\pm0.4}$
 \\ \bottomrule
\end{tabular}%
}
\end{center}
\end{table*}

\section{Further Analysis}

In this section, we provide further analysis on the training dynamics of Pi-DUAL, the distribution of the learned gates and several ablations on our method. Overall, we show that Pi-DUAL behaves as 
expected from its design,
and that all pieces of its architecture contribute to its good performance. 

\subsection{Training Dynamics}\label{sec:dynamics}

To verify that Pi-DUAL effectively decouples the learning paths of samples with correct and wrong labels, we study the training dynamics of the prediction and noise networks on each of these sets of samples, in comparison to the training dynamics of cross-entropy baseline. We observe in Fig.~\ref{fig:train_curve} that the prediction network of Pi-DUAL mostly fits the correct labels, as its training accuracy on samples with wrong labels is always very low on 
all datasets\footnote{Results for other datasets are shown in Appendix~\ref{app:training_dynamics}.}. Meanwhile, the noise network shows the opposite behavior and mostly fits the wrong labels on CIFAR-10H and CIFAR-100N. Interestingly, we observe that the noise network does not fit any samples on ImageNet-PI. We attribute this behavior to the fact that ImageNet has more than a million samples and 1000 classes, so fitting the noise is very hard. Indeed, as shown on the bottom row of Fig.~\ref{fig:train_curve}, the cross-entropy baseline also ignores the samples with wrong labels. However, the cross-entropy baseline has lower training accuracy on the correct labels than Pi-DUAL as it cannot effectively separate the two distributions, and therefore achieves worse test accuracy.\looseness=-1

In all datasets, we see that the test accuracy of Pi-DUAL grows gradually and steadily with training and that overfitting to the wrong labels does not hurt its performance as these are mostly fit by the noise network $\epsilon_{\bm\phi}$. Meanwhile, we observe that on CIFAR-10H and CIFAR-100N, the test accuracy of the cross-entropy baseline starts degrading as the accuracy on samples with wrong labels starts to grow. This is a clear sign that Pi-DUAL effectively leverages the PI to learn shortcuts that protect the feature extraction of $f_{\bm{\theta}}$ and therefore does not require to use early-stopping to achieve its best results.\looseness=-1

\subsection{Analysis of the Gating Network Predictions}\label{sec:gates}

In our model, the gating network $\gamma_\psi$ is tasked with learning the binary indicator signal $\gamma$, which tells whether a sample belongs to $\mathcal{D}_{\text{correct}}$ or $\mathcal{D}_{\text{wrong}}$. To show that the model works as intended, we plot in Fig.~\ref{fig:gate_distribution} the distribution of $\gamma_{\psi}(\bm a)$ separately for samples with correct and wrong labels after training on different datasets\footnote{Results for other datasets are shown in Appendix~\ref{app:gate_distribution}.}. As expected, in the two datasets with high-quality PI -- CIFAR-10H and ImageNet-PI -- the gate distribution achieves a separation between the two distributions (cf. Tab.~\ref{table:auc_number}). And even in the case of CIFAR-100N, where the PI is not very informative, the gate output still separates a big portion of the wrong labels.\looseness=-1

To give a better intuition of what Pi-DUAL learns, we provide some visual examples of both success and failures cases of the gating network when training on ImageNet-PI (high-noise). As shown in Fig.~\ref{fig:visual_example}, the gating network can often detect blatantly wrong annotations which are further corrected by the prediction network. Interestingly, we observe that in the few cases where the gating network makes a mistake, the predicted clean label is not so far from what many humans would suggest -- like in the crane picture on the bottom right which is recognized by Pi-DUAL as a fire truck.\looseness=-1

\subsection{Ablation Studies}\label{sec:ablation}

We finally present various ablation studies analysing the contributions of different components of Pi-DUAL in Tab.~\ref{table:ablation_structure}.

\textbf{Architecture ablation.} Pi-DUAL gives $\va$ as input to both the noise network $\epsilon_{\bm\phi}$ and the gating network $\gamma_{\bm\psi}$. As shown in Tab.~\ref{table:ablation_structure}, removing either of the two elements from the architecture generally results in lower performance gains than with the full architecture. Interestingly, on ImageNet-PI, the noise network does not seem to be critical. We attribute this behavior to the fact that, on these datasets, Pi-DUAL does not need to overfit to the noisy labels to achieve good performance (cf. Fig.~\ref{fig:train_curve}). Indeed, just using the gating mechanism to toggle on-off the fitting of the noisy labels seems sufficient to achieve good performance in a dataset with so many classes. We provide more ablation studies on the model architecture in Appendix \ref{app:ablation_structure}, including the model backbone for the prediction network and the network structure for the noise and gating network. Pi-DUAL delivers consistent, high performance with different architectures. 
\looseness=-1

\footnotetext[3]{The training accuracy for the noise network on ImageNet-PI (high noise) is 0.07\% and 0.1\% respectively for correct and wrong labels.}
\textbf{Gating in probability space.} In Sec.~\ref{sec:method} we chose to parameterize Pi-DUAL in the logit space. An alternative is to parameterize the gating mechanism in the probability space. However, although the probabilistic version of Pi-DUAL performs better than the cross-entopy baseline in most cases, it underperforms compared to the logit space version.

\textbf{Performance without PI.} We argued before that Pi-DUAL performs better on datasets with high-quality PI as this permits to wield the most power from its structure. For completeness, we now test the performance of Pi-DUAL without access to dataset-specific PI features. That is, having only access to the random PI sample-identifier proposed by \citet{ortiz2023does}. We see that without access to PI features, Pi-DUAL still can  perform better than the cross-entropy baseline, but its performance deteriorates significantly, i.e., having access to good PI is fundamental for Pi-DUAL's success.\looseness=-1

\section{Conclusion}
\label{sec:conclusion}

In this paper, we have presented Pi-DUAL, a new method that utilizes PI to combat label noise by introducing a dual network structure designed to model the generative process of the noisy annotations. Experimental results have demonstrated the effectiveness of Pi-DUAL in learning to both fit the clean label distribution and detect noisy samples. Pi-DUAL sets a new state-of-the-art accuracy in datasets with high-quality PI features. We have performed extensive ablation studies and thorough analysis, both empirical and theoretical, to provide insights into how Pi-DUAL works. Importantly, Pi-DUAL is very easy to implement and can be plugged into any training pipeline. Unlike competing approaches, it gracefully scales up to datasets with millions of examples and thousands of classes.  Moving forward, it will be interesting to study extensions of Pi-DUAL that can also tackle other problems with PI beyond supervised classification.%

\section*{Impact Statement}
This paper presents work whose goal is to advance the field of Machine Learning. 

Overall, we do not see any special ethical concerns stemming directly from our work. In particular we note that although annotator IDs are part of the PI used in our experiments, none of our results require the use of personally identifiable annotator IDs. In fact, cryptographically safe IDs in the form of hashes work equally well as PI. In this regard, we do not think that there are serious concerns about possible identity leakages stemming from the proposed framework if the proper anonymization protocols are followed.\looseness=-1

\section*{Acknowledgments}
We thank the anonymous reviewers for helpful feedback and comments. 

\bibliography{icml2024}

\begin{thebibliography}{43}
\providecommand{\natexlab}[1]{#1}
\providecommand{\url}[1]{\texttt{#1}}
\expandafter\ifx\csname urlstyle\endcsname\relax
  \providecommand{\doi}[1]{doi: #1}\else
  \providecommand{\doi}{doi: \begingroup \urlstyle{rm}\Url}\fi

\bibitem[Albert et~al.(2023)Albert, Arazo, Krishna, O’Connor, and McGuinness]{albert2023your}
Albert, P., Arazo, E., Krishna, T., O’Connor, N.~E., and McGuinness, K.
\newblock Is your noise correction noisy? pls: Robustness to label noise with two stage detection.
\newblock In \emph{IEEE Winter Conference on Applications of Computer Vision (WACV)}, 2023.

\bibitem[Arpit et~al.(2017)Arpit, Jastrz{\k{e}}bski, Ballas, Krueger, Bengio, Kanwal, Maharaj, Fischer, Courville, Bengio, et~al.]{arpit2017closer}
Arpit, D., Jastrz{\k{e}}bski, S., Ballas, N., Krueger, D., Bengio, E., Kanwal, M.~S., Maharaj, T., Fischer, A., Courville, A., Bengio, Y., et~al.
\newblock A closer look at memorization in deep networks.
\newblock In \emph{International Conference on Machine Learning (ICML)}, 2017.

\bibitem[Bach(2021)]{bach2021learning}
Bach, F.
\newblock \emph{Learning Theory from First Principles}.
\newblock (draft), 2021.

\bibitem[Bai et~al.(2021)Bai, Yang, Han, Yang, Li, Mao, Niu, and Liu]{bai2021understanding}
Bai, Y., Yang, E., Han, B., Yang, Y., Li, J., Mao, Y., Niu, G., and Liu, T.
\newblock Understanding and improving early stopping for learning with noisy labels.
\newblock In \emph{Advances in Neural Information Processing Systems (NeurIPS)}, 2021.

\bibitem[Baldock et~al.(2021)Baldock, Maennel, and Neyshabur]{baldock_2021}
Baldock, R. J.~N., Maennel, H., and Neyshabur, B.
\newblock Deep learning through the lens of example difficulty.
\newblock In \emph{Advances in Neural Information Processing Systems (NeurIPS)}, 2021.

\bibitem[Berthelot et~al.(2019)Berthelot, Carlini, Goodfellow, Papernot, Oliver, and Raffel]{berthelot2019mixmatch}
Berthelot, D., Carlini, N., Goodfellow, I., Papernot, N., Oliver, A., and Raffel, C.~A.
\newblock Mixmatch: A holistic approach to semi-supervised learning.
\newblock 2019.

\bibitem[Cheng et~al.(2020)Cheng, Zhu, Li, Gong, Sun, and Liu]{cheng2020learning}
Cheng, H., Zhu, Z., Li, X., Gong, Y., Sun, X., and Liu, Y.
\newblock Learning with instance-dependent label noise: A sample sieve approach.
\newblock \emph{arXiv preprint arXiv:2010.02347}, 2020.

\bibitem[Collier et~al.(2021)Collier, Mustafa, Kokiopoulou, Jenatton, and Berent]{het}
Collier, M., Mustafa, B., Kokiopoulou, E., Jenatton, R., and Berent, J.
\newblock Correlated input-dependent label noise in large-scale image classification.
\newblock In \emph{{IEEE} Conference on Computer Vision and Pattern Recognition ({CVPR})}, 2021.

\bibitem[Collier et~al.(2022)Collier, Jenatton, Kokiopoulou, and Berent]{collier2022transfer}
Collier, M., Jenatton, R., Kokiopoulou, E., and Berent, J.
\newblock Transfer and marginalize: Explaining away label noise with privileged information.
\newblock In \emph{International Conference on Machine Learning (ICML)}, 2022.

\bibitem[Collier et~al.(2023)Collier, Jenatton, Mustafa, Houlsby, Berent, and Kokiopoulou]{collier2023hetxl}
Collier, M., Jenatton, R., Mustafa, B., Houlsby, N., Berent, J., and Kokiopoulou, E.
\newblock Massively scaling heteroscedastic classifiers.
\newblock In \emph{International Conference on Learning Representations (ICLR)}, 2023.

\bibitem[Deng et~al.(2009)Deng, Dong, Socher, Li, Li, and Fei-Fei]{deng2009imagenet}
Deng, J., Dong, W., Socher, R., Li, L.-J., Li, K., and Fei-Fei, L.
\newblock Imagenet: A large-scale hierarchical image database.
\newblock In \emph{IEEE Conference on Computer Vision and Pattern Recognition (CVPR)}, 2009.

\bibitem[Han et~al.(2018)Han, Yao, Yu, Niu, Xu, Hu, Tsang, and Sugiyama]{han2018co}
Han, B., Yao, Q., Yu, X., Niu, G., Xu, M., Hu, W., Tsang, I., and Sugiyama, M.
\newblock Co-teaching: Robust training of deep neural networks with extremely noisy labels.
\newblock 2018.

\bibitem[Hern{\'a}ndez-Lobato et~al.(2014)Hern{\'a}ndez-Lobato, Sharmanska, Kersting, Lampert, and Quadrianto]{hernandez2014mind}
Hern{\'a}ndez-Lobato, D., Sharmanska, V., Kersting, K., Lampert, C.~H., and Quadrianto, N.
\newblock Mind the nuisance: Gaussian process classification using privileged noise.
\newblock In \emph{Advances in Neural Information Processing Systems (NeurIPS}, 2014.

\bibitem[Jiang et~al.(2018)Jiang, Zhou, Leung, Li, and Fei-Fei]{jiang2018mentornet}
Jiang, L., Zhou, Z., Leung, T., Li, L.-J., and Fei-Fei, L.
\newblock Mentornet: Learning data-driven curriculum for very deep neural networks on corrupted labels.
\newblock In \emph{International Conference on Machine Learning (ICML)}, 2018.

\bibitem[Krizhevsky et~al.(2009)Krizhevsky, Hinton, et~al.]{krizhevsky2009learning}
Krizhevsky, A., Hinton, G., et~al.
\newblock Learning multiple layers of features from tiny images.
\newblock 2009.

\bibitem[Li et~al.(2020{\natexlab{a}})Li, Socher, and Hoi]{li2020dividemix}
Li, J., Socher, R., and Hoi, S.~C.
\newblock Dividemix: Learning with noisy labels as semi-supervised learning.
\newblock \emph{arXiv preprint arXiv:2002.07394}, 2020{\natexlab{a}}.

\bibitem[Li et~al.(2020{\natexlab{b}})Li, Socher, and Hoi]{dividemix}
Li, J., Socher, R., and Hoi, S. C.~H.
\newblock Dividemix: Learning with noisy labels as semi-supervised learning.
\newblock In \emph{International Conference on Learning Representations (ICLR)}, 2020{\natexlab{b}}.

\bibitem[Liu et~al.(2020)Liu, Niles-Weed, Razavian, and Fernandez-Granda]{liu2020early}
Liu, S., Niles-Weed, J., Razavian, N., and Fernandez-Granda, C.
\newblock Early-learning regularization prevents memorization of noisy labels.
\newblock In \emph{Advances in Neural Information Processing Systems (NeurIPS)}, 2020.

\bibitem[Liu et~al.(2022)Liu, Zhu, Qu, and You]{liu2022robust}
Liu, S., Zhu, Z., Qu, Q., and You, C.
\newblock Robust training under label noise by over-parameterization.
\newblock In \emph{International Conference on Machine Learning (ICML)}, 2022.

\bibitem[Lopez-Paz et~al.(2015)Lopez-Paz, Bottou, Sch{\"o}lkopf, and Vapnik]{lopez2015unifying}
Lopez-Paz, D., Bottou, L., Sch{\"o}lkopf, B., and Vapnik, V.
\newblock Unifying distillation and privileged information.
\newblock \emph{arXiv preprint arXiv:1511.03643}, 2015.

\bibitem[Loshchilov \& Hutter(2016)Loshchilov and Hutter]{loshchilov2016sgdr}
Loshchilov, I. and Hutter, F.
\newblock Sgdr: Stochastic gradient descent with warm restarts.
\newblock \emph{arXiv preprint arXiv:1608.03983}, 2016.

\bibitem[Maennel et~al.(2020)Maennel, Alabdulmohsin, Tolstikhin, Baldock, Bousquet, Gelly, and Keysers]{maennel_2020}
Maennel, H., Alabdulmohsin, I.~M., Tolstikhin, I.~O., Baldock, R. J.~N., Bousquet, O., Gelly, S., and Keysers, D.
\newblock What do neural networks learn when trained with random labels?
\newblock In \emph{Advances in Neural Information Processing Systems (NeurIPS)}, 2020.

\bibitem[Menon et~al.(2020)Menon, Rawat, Reddi, and Kumar]{menon2020can}
Menon, A.~K., Rawat, A.~S., Reddi, S.~J., and Kumar, S.
\newblock Can gradient clipping mitigate label noise?
\newblock In \emph{International Conference on Learning Representations (ICLR)}, 2020.

\bibitem[Nado et~al.(2021)Nado, Band, Collier, Djolonga, Dusenberry, Farquhar, Feng, Filos, Havasi, Jenatton, et~al.]{nado2021uncertainty}
Nado, Z., Band, N., Collier, M., Djolonga, J., Dusenberry, M.~W., Farquhar, S., Feng, Q., Filos, A., Havasi, M., Jenatton, R., et~al.
\newblock Uncertainty baselines: Benchmarks for uncertainty \& robustness in deep learning.
\newblock \emph{arXiv preprint arXiv:2106.04015}, 2021.

\bibitem[Ortiz-Jimenez et~al.(2023)Ortiz-Jimenez, Collier, Nawalgaria, D'Amour, Berent, Jenatton, and Kokiopoulou]{ortiz2023does}
Ortiz-Jimenez, G., Collier, M., Nawalgaria, A., D'Amour, A., Berent, J., Jenatton, R., and Kokiopoulou, E.
\newblock When does privileged information explain away label noise?
\newblock In \emph{International Conference on Machine Learning (ICML)}, 2023.

\bibitem[Patrini et~al.(2017)Patrini, Rozza, Krishna~Menon, Nock, and Qu]{patrini2017making}
Patrini, G., Rozza, A., Krishna~Menon, A., Nock, R., and Qu, L.
\newblock Making deep neural networks robust to label noise: A loss correction approach.
\newblock In \emph{IEEE Conference on Computer Vision and Pattern Recognition (CVPR)}, 2017.

\bibitem[Peterson et~al.(2019)Peterson, Battleday, Griffiths, and Russakovsky]{cifarh}
Peterson, J.~C., Battleday, R.~M., Griffiths, T.~L., and Russakovsky, O.
\newblock Human uncertainty makes classification more robust.
\newblock In \emph{{IEEE} International Conference on Computer Vision ({ICCV})}, 2019.

\bibitem[Riquelme et~al.(2021)Riquelme, Puigcerver, Mustafa, Neumann, Jenatton, Susano~Pinto, Keysers, and Houlsby]{riquelme2021scaling}
Riquelme, C., Puigcerver, J., Mustafa, B., Neumann, M., Jenatton, R., Susano~Pinto, A., Keysers, D., and Houlsby, N.
\newblock Scaling vision with sparse mixture of experts.
\newblock \emph{Advances in Neural Information Processing Systems}, 34:\penalty0 8583--8595, 2021.

\bibitem[Shazeer et~al.(2017)Shazeer, Mirhoseini, Maziarz, Davis, Le, Hinton, and Dean]{shazeer2017outrageously}
Shazeer, N., Mirhoseini, A., Maziarz, K., Davis, A., Le, Q., Hinton, G., and Dean, J.
\newblock Outrageously large neural networks: The sparsely-gated mixture-of-experts layer.
\newblock In \emph{International Conference on Learning Representations (ICLR)}, 2017.

\bibitem[Sheng et~al.(2008)Sheng, Provost, and Ipeirotis]{sheng_2008}
Sheng, V.~S., Provost, F., and Ipeirotis, P.~G.
\newblock Get another label? improving data quality and data mining using multiple, noisy labelers.
\newblock In \emph{ACM SIGKDD International Conference on Knowledge Discovery and Data Mining (KDD)}, 2008.

\bibitem[Snow et~al.(2008)Snow, O{'}Connor, Jurafsky, and Ng]{snow-etal-2008-cheap}
Snow, R., O{'}Connor, B., Jurafsky, D., and Ng, A.
\newblock Cheap and fast {--} but is it good? evaluating non-expert annotations for natural language tasks.
\newblock In \emph{Conference on Empirical Methods in Natural Language Processing (EMNLP)}, 2008.

\bibitem[Sohn et~al.(2020)Sohn, Berthelot, Carlini, Zhang, Zhang, Raffel, Cubuk, Kurakin, and Li]{sohn2020fixmatch}
Sohn, K., Berthelot, D., Carlini, N., Zhang, Z., Zhang, H., Raffel, C.~A., Cubuk, E.~D., Kurakin, A., and Li, C.-L.
\newblock Fixmatch: Simplifying semi-supervised learning with consistency and confidence, 2020.

\bibitem[Song et~al.(2022)Song, Kim, Park, Shin, and Lee]{song2022learning}
Song, H., Kim, M., Park, D., Shin, Y., and Lee, J.-G.
\newblock Learning from noisy labels with deep neural networks: A survey.
\newblock \emph{IEEE Transactions on Neural Networks and Learning Systems}, 2022.

\bibitem[Szegedy et~al.(2016)Szegedy, Vanhoucke, Ioffe, Shlens, and Wojna]{szegedy2016rethinking}
Szegedy, C., Vanhoucke, V., Ioffe, S., Shlens, J., and Wojna, Z.
\newblock Rethinking the inception architecture for computer vision.
\newblock In \emph{IEEE Conference on Computer Vision and Pattern Recognition (CVPR)}, 2016.

\bibitem[Tanaka et~al.(2018)Tanaka, Ikami, Yamasaki, and Aizawa]{tanaka2018joint}
Tanaka, D., Ikami, D., Yamasaki, T., and Aizawa, K.
\newblock Joint optimization framework for learning with noisy labels.
\newblock In \emph{IEEE Conference on Computer Vision and Pattern Recognition (CVPR)}, 2018.

\bibitem[Vapnik \& Vashist(2009)Vapnik and Vashist]{lupi_vapnik}
Vapnik, V. and Vashist, A.
\newblock A new learning paradigm: Learning using privileged information.
\newblock \emph{Neural Networks}, 2009.

\bibitem[Wang et~al.(2023)Wang, Xiao, Dong, Feng, and Zhao]{wang2022promix}
Wang, H., Xiao, R., Dong, Y., Feng, L., and Zhao, J.
\newblock {ProMix: combating label noise via maximizing clean sample utility}.
\newblock In \emph{International Joint Conferences on Artificial Intelligence (IJCAI)}, 2023.

\bibitem[Wei et~al.(2021)Wei, Zhu, Cheng, Liu, Niu, and Liu]{wei2021learning}
Wei, J., Zhu, Z., Cheng, H., Liu, T., Niu, G., and Liu, Y.
\newblock Learning with noisy labels revisited: A study using real-world human annotations.
\newblock \emph{arXiv preprint arXiv:2110.12088}, 2021.

\bibitem[Wei et~al.(2022)Wei, Zhu, Cheng, Liu, Niu, and Liu]{cifarn}
Wei, J., Zhu, Z., Cheng, H., Liu, T., Niu, G., and Liu, Y.
\newblock Learning with noisy labels revisited: A study using real-world human annotations.
\newblock In \emph{International Conference on Learning Representations (ICLR)}, 2022.

\bibitem[Xie et~al.(2020)Xie, Dai, Hovy, Luong, and Le]{xie2020unsupervised}
Xie, Q., Dai, Z., Hovy, E., Luong, T., and Le, Q.
\newblock Unsupervised data augmentation for consistency training.
\newblock In \emph{Advances in Neural Information Processing Systems (NeurIPS)}, 2020.

\bibitem[Yi \& Wu(2019)Yi and Wu]{yi2019probabilistic}
Yi, K. and Wu, J.
\newblock Probabilistic end-to-end noise correction for learning with noisy labels.
\newblock In \emph{IEEE Conference on Computer Vision and Pattern Recognition (CVPR)}, 2019.

\bibitem[Yu et~al.(2019)Yu, Han, Yao, Niu, Tsang, and Sugiyama]{yu2019does}
Yu, X., Han, B., Yao, J., Niu, G., Tsang, I., and Sugiyama, M.
\newblock How does disagreement help generalization against label corruption?
\newblock In \emph{International Conference on Machine Learning (ICML)}, 2019.

\bibitem[Zhang et~al.(2017)Zhang, Bengio, Hardt, Recht, and Vinyals]{ZhangBengioHardtRechtVinyals.17}
Zhang, C., Bengio, S., Hardt, M., Recht, B., and Vinyals, O.
\newblock Understanding deep learning requires rethinking generalization.
\newblock In \emph{International Conference on Learning Representations (ICLR)}, 2017.

\end{thebibliography}
\bibliographystyle{icml2024}

\newpage
\appendix

\onecolumn

\section{Theoretical Insights: Risk Analysis}\label{app:risk_analysis}

\paragraph{Model and notations.} We assume the following regression setting
\begin{equation*}
    \vy =
\begin{bmatrix}
\vy_1  \\
\vy_2
\end{bmatrix}
=
\begin{bmatrix}
\mX_1 \vw^\star  \\
\vzero
\end{bmatrix} +
\begin{bmatrix}
\vzero  \\
\mA_2 \vv^\star
\end{bmatrix} +
\begin{bmatrix}
\rvvarepsilon_1  \\
\rvvarepsilon_2
\end{bmatrix} \in \R^n 
\end{equation*}
where we have $n=n_1+n_2$ observations such that
\begin{itemize}
    \item $\vy_1 = \mX_1 \vw^\star + \rvvarepsilon_1 \in \R^{n_1}$ with $\mX_1 \in \R^{n_1 \times d}$ and $\rvvarepsilon_1 \sim \gN(\vzero, \sigma^2\mI )$,
    \item $\vy_2 = \mA_2 \vv^\star + \rvvarepsilon_2 \in \R^{n_2}$ with $\mA_2 \in \R^{n_2 \times m}$ and $\rvvarepsilon_2 \sim \gN(\vzero, \sigma^2\mI )$.
\end{itemize}
The vector $\vy_1$ corresponds to the clean targets that depend on the features $\mX_1$ while $\vy_2$ corresponds to the noisy targets that are explained by the privileged information (PI) represented by $\mA_2$.

We use the matrix forms $\mX = [\mX_1, \mX_2] \in \R^{n \times d}$, $\mA = [\mA_1, \mA_2] \in \R^{n \times m}$ and $\rvvarepsilon = [\rvvarepsilon_1, \rvvarepsilon_2] \in \R^n$. Moreover, we consider the diagonal mask matrix $\rvgamma^\star \in \{0, 1\}^{n \times n}$ such that
\begin{equation*}
    \rvgamma^* \mX =
\begin{bmatrix}
\mX_1  \\
\vzero
\end{bmatrix}
\ \text{and}\ 
(\mI - \rvgamma^*) \mA =
\begin{bmatrix}
\vzero \\
\mA_2
\end{bmatrix}.
\end{equation*}
We list below some notation that we will repeatedly use
\begin{itemize}
    \item The covariance matrices $\mQ = \mX^\top \mX$ and $\mQ_1 = \mX_1^\top \mX_1$
    \item The difference between the contributions of the standard features and the PI features
    \begin{equation*}
        \rvdelta^\star = \mX \vw^\star - \mA \vv^\star \in \R^n
    \end{equation*}
    \item The orthogonal projector onto the span of the columns of $\mX$:
    \begin{equation*}
        \mPi_x = \mX (\mX^\top \mX)^{-1} \mX^\top \in \R^{n \times n}.
    \end{equation*}
    \item For any diagonal mask matrix $\rvgamma \in \{0, 1\}^{n \times n}$, we define the diagonal matrix that records the differences with respect to the reference $\rvgamma^\star$ 
    \begin{equation*}
        \mDelta_\gamma = \rvgamma^\star - \rvgamma \in \{-1, 0, 1\}^{n \times n}.
    \end{equation*}
\end{itemize}
The rest of our exposition follows the structure of~\citet{collier2022transfer}.

\subsection{Definition of the Risk}

To compare different predictors, we will consider their \textit{risks}, that is, their ability to generalize. We focus on the \textit{fixed} design analysis~\citep{bach2021learning}, i.e., we study the errors only due to resampling the additive noise $\varepsilon$.
In our context, we are more specifically interested in \textit{the performance of the predictors on the clean targets} (with predictors having been trained on both clean and noisy targets). 

Formally, given a predictor $\vtheta$ based on the training quantities $(\mX, \mA, \rvvarepsilon)$, we consider
\begin{equation*}
\vy_1' = \mX_1 \vw^\star + \rvvarepsilon_1'
\end{equation*}
where the prime $'$ is to show the difference with the training quantities without prime, and we define the risk of $\vtheta$ as
\begin{equation}
    \gR(\vtheta) = \E_{\varepsilon'_1 \sim p(\varepsilon'_1)}\bigg\{\frac{1}{n_1} \|\vy_1' - \mX_1 \vtheta \|^2 \bigg\}.
\end{equation}
A simple expansion of the square with $\E_{\varepsilon'}[\|\rvvarepsilon'_1\|^2] = n_1 \sigma^2$ leads to the standard expression 
\begin{equation}\label{eq:expanded_risk}
    \gR(\vtheta) = \frac{1}{n_1} \|\mX_1(\vtheta - \vw^\star) \|^2 + \sigma^2
    = \frac{1}{n_1} \|\rvgamma^\star \mX (\vtheta - \vw^\star) \|^2 + \sigma^2.
\end{equation}
To obtain the final expression of the risk, we eventually take a second expectation $\E_{\varepsilon \sim p(\varepsilon)}[\gR(\vtheta)]$ with respect to the training quantity $\rvvarepsilon$~\citep{bach2021learning}.

\subsection{Main Result}

We state below our main result and discuss its implications.

\begin{proposition}
Consider some diagonal mask matrix $\rvgamma \in \{0, 1\}^{n\times n}$ and the masked versions of $\mX$ and $\mA$ which we refer to as $\bar{\mX} = \rvgamma \mX$ and $\bar{\mA} = (\mI - \rvgamma) \mA$.

Let us assume that $\mQ \in \R^{d \times d}$, $\bar{\mX}^\top \bar{\mX}  \in \R^{d \times d}$, $\bar{\mA}^\top \bar{\mA}  \in \R^{m \times m}$ and
\begin{equation}
    \begin{bmatrix}
    \bar{\mX}^\top \bar{\mX}\ \ \ \ \bar{\mX}^\top \bar{\mA} \\ \vspace{-0.1cm}\\
    \bar{\mA}^\top \bar{\mX}\ \ \ \ \bar{\mA}^\top \bar{\mA}
    \end{bmatrix} \in \R^{(d+m) \times (d+m)}\label{eq:X_A_covariance}
\end{equation}
are all invertible.
Let us define by $\vw_0$ the ordinary-least-squares predictor (see Eq. (\ref{eq:ols_problem})). Similarly, let us define by $\vw_1$ the Pi-DUAL predictor, using $\rvgamma$ as (pre-defined) gates (see Eq. (\ref{eq:pi_AG_problem})).

It holds that the risk $\E[\gR(\vw_0)]$ of $\vw_0$ is larger than the risk $\E[\gR(\vw_1)]$ of $\vw_1$ if and only if
\begin{equation}
    \bias{\| \rvgamma^\star \mPi_x (\mI - \rvgamma^*)\rvdelta^\star\|^2} + 
    \variance{\sigma^2 \mathrm{tr}(\mQ^{-1} \mQ_1)}
    > 
    \bias{ \| \rvgamma^\star \mX \bar{\mH} \mDelta_\rvgamma \rvdelta^\star \|^2} + 
    \variance{\sigma^2 \mathrm{tr}(\bar{\mQ}_a^{-1} \mQ_1)}\label{eq:risk_comparison}
\end{equation}
where the matrices $ \bar{\mH}$ and $\bar{\mQ}_a$ are defined in Section~\ref{sec:proof_pi_AG}.
\end{proposition}
The proofs of the risk expressions can be found in Sections \ref{sec:proof_ols} and \ref{sec:proof_pi_AG}.

\subsubsection{Discussion}
The condition in Eq.~(\ref{eq:risk_comparison}) brings into play the \bias{bias terms} and the \variance{variance terms} of the risks of $\vw_0$ and $\vw_1$.\looseness=-1

As intuitively expected, the \variance{variance term} corresponding to $\vw_0$ is smaller than that of $\vw_1$. Indeed, Pi-DUAL requires to learn more parameters (both $\vw$ and $\vv$) than in the case of the standard ordinary least squares.
More precisely, if the spans of the columns $\bar{\mA}$ and $\bar{\mX}$ are close to be orthogonal to each other (as suggested by the invertibility condition for Eq. (\ref{eq:X_A_covariance})), we approximately have
\begin{equation*}
    \text{tr}(\mQ^{-1} \mQ_1) \approx d\frac{n_1}{n} 
    <
    \text{tr}(\bar{\mQ}_a^{-1} \mQ_1) \approx d\frac{n_1}{\bar{n}_1}
\end{equation*}
where $\bar{n}_1 = \vone^\top \rvgamma \vone$ stands for the number of examples selected by the gate $\rvgamma$ (with $\bar{n}_1 < n$).

When looking at the \bias{bias terms}, we see how Pi-DUAL can compensate for a larger variance term to achieve a lower risk overall.
We first recall the definition of $\rvdelta^\star=\mX \vw^\star - \mA \vv^\star$ that computes the difference between the contributions of the standard features $\mX$ and the PI features $\mA$. If the level of noise explained by $\mA_2$ has a large contribution compared with the signal from $\mX_2$, the second part $\rvdelta^\star_2$ of $\rvdelta^\star$ can contain large entries.
While $\vw_0$ has a bias term scaling with $\gO((\mI - \rvgamma^*)\rvdelta^\star)$---that is, proportional to the number $n_2$ of noisy examples captured by $\rvdelta^\star_2$---we can observe that $\vw_1$ has a more robust scaling. Indeed, it depends on $\gO(\mDelta_\rvgamma \rvdelta^\star)$ that only scales with the number of disagreements between the reference gate $\rvgamma^\star$ and that used for training $\rvgamma$.

\subsection{Proof: Risk of Ordinary Least Squares}\label{sec:proof_ols}
We assume that $\mQ$ is invertible.
We focus on the solution of
\begin{equation}
\min_{\vw \in \R^d} \frac{1}{2n} \| \vy - \mX \vw \|^2\label{eq:ols_problem}
\end{equation}
that is given by 
\begin{eqnarray}
    \vw_0 &=& \mQ^{-1} \mX^\top \vy \nonumber \\
    &=& \mQ^{-1} \mX^\top (\rvgamma^* \mX \vw^\star + (\mI - \rvgamma^*) \mA \vv^\star + \rvvarepsilon) \nonumber \\
    &=& \mQ^{-1} \mX^\top (-(\mI - \rvgamma^*)\rvdelta^\star + \mX \vw^\star  + \rvvarepsilon) \nonumber \\
    &=& - \mQ^{-1} \mX^\top (\mI - \rvgamma^*)\rvdelta^\star + \vw^\star + \mQ^{-1} \mX^\top \rvvarepsilon. \nonumber
\end{eqnarray}
Plugging into Eq.~(\ref{eq:expanded_risk}), we obtain
\begin{equation*}
\gR(\vw_0) = \frac{1}{n_1} \| \rvgamma^\star \mPi_x (\mI - \rvgamma^*)\rvdelta^\star  - \rvgamma^\star \mPi_x \rvvarepsilon\|^2 + \sigma^2.
\end{equation*}
Expanding the square and using that $\text{tr}(\rvgamma^* \mPi_x (\rvgamma^* \mPi_x)^\top)=\text{tr}(\rvgamma^* \mPi_x)=\text{tr}(\mQ^{-1} \mQ_1)$, the final risk expression is
\begin{eqnarray}
    \E[\gR(\vw_0)] & = & \frac{1}{n_1} \| \rvgamma^\star \mPi_x (\mI - \rvgamma^*)\rvdelta^\star \|^2 + \frac{1}{n_1} \E[\|\rvgamma^\star \mPi_x \rvvarepsilon \|^2] + \sigma^2 \nonumber \\
    & = & \frac{1}{n_1} \| \rvgamma^\star \mPi_x (\mI - \rvgamma^*)\rvdelta^\star\|^2 + \frac{\sigma^2}{n_1} \text{tr}(\mQ^{-1} \mQ_1)+ \sigma^2. \label{eq:risk_ols}
\end{eqnarray}

\subsection{Proof: Risk of Pi-DUAL}\label{sec:proof_pi_AG}
We focus on the solution of
\begin{equation}
\min_{\vw\in\R^d, \vv\in\R^m} \frac{1}{2n} \| \vy - (\rvgamma \mX \vw + (\mI - \rvgamma) \mA \vv) \|^2\label{eq:pi_AG_problem}
\end{equation}
to construct an estimator. 
Here, $\rvgamma$ refers to a diagonal mask matrix of size $n \times n$ which we use as (pre-defined) gates for Pi-DUAL.
We introduce the notations:
\begin{itemize}
    \item The masked versions of $\mX$ and $\mA$: $\bar{\mX} = \rvgamma \mX$ and $\bar{\mA} = (\mI - \rvgamma) \mA$
    \item The projector onto the span of the columns of $\bar{\mA}$:
    \begin{equation*}
        \bar{\mPi}_a = \bar{\mA} (\bar{\mA}^\top \bar{\mA})^{-1} \bar{\mA}^\top \in \R^{n \times n}
    \end{equation*}
    \item The projection $\bar{\mX}_a = (\mI - \bar{\mPi}_a) \bar{\mX} \in \R^{n \times d}$ of $\bar{\mX}$ onto the orthogonal of the span of the columns of $\bar{\mA}$, and the matrices
    \begin{equation*}
        \bar{\mH} = (\bar{\mX}_a^\top \bar{\mX}_a)^{-1} \bar{\mX}_a^\top \in \R^{d \times n}
        \ \text{and}\ \bar{\mQ}_a = \bar{\mX}_a^\top \bar{\mX}_a \in \R^{d \times d}.
    \end{equation*}
\end{itemize}
We can reuse Lemma~I.3 from~\citet{collier2022transfer}, with $(\bar{\mX}, \bar{\mA})$ in lieu of $(\mX, \mA)$. The solution of $\vw$ is thus given by
\begin{eqnarray}
    \vw_1 &=& \bar{\mH} \vy \nonumber \\
    &=& \bar{\mH} (\rvgamma^* \mX \vw^\star + (\mI - \rvgamma^*) \mA \vv^\star + \rvvarepsilon) \nonumber \\
    &=& \bar{\mH} ((\mDelta_\rvgamma + \rvgamma) \mX \vw^\star + (\mI - (\mDelta_\rvgamma + \rvgamma)) \mA \vv^\star + \rvvarepsilon) \nonumber \\
    &=& \bar{\mH} (\mDelta_\rvgamma \rvdelta^\star + \bar{\mX} \vw^\star + \bar{\mA} \vv^\star + \rvvarepsilon) \nonumber \\
    &=& \bar{\mH} \mDelta_\rvgamma \rvdelta^\star +  \vw^\star + \vzero + \bar{\mH} \rvvarepsilon. \nonumber
\end{eqnarray}
where in the last line, we have used that $\bar{\mH} \bar{\mX} = (\bar{\mX}_a^\top \bar{\mX}_a)^{-1} \bar{\mX}_a^\top \bar{\mX}_a=\mI$ (because $\mI - \bar{\mPi}_a = (\mI - \bar{\mPi}_a)^2$) and $(\mI - \bar{\mPi}_a) \bar{\mA} = \vzero$.

Plugging into Eq.~(\ref{eq:expanded_risk}), we obtain
\begin{equation*}
\gR(\vw_1) = \frac{1}{n_1} \| \rvgamma^\star \mX \bar{\mH} \mDelta_\rvgamma \rvdelta^\star +  \rvgamma^\star \mX \bar{\mH} \rvvarepsilon\|^2 + \sigma^2.
\end{equation*}
Expanding the square and using that $\text{tr}(\rvgamma^\star \mX \bar{\mH} (\rvgamma^\star \mX \bar{\mH})^\top)=\text{tr}(\rvgamma^\star \mX \bar{\mQ}_a^{-1} \mX^\top \rvgamma^\star)=\text{tr}(\bar{\mQ}_a^{-1} \mQ_1)$, the final risk expression is
\begin{eqnarray}
    \E[\gR(\vw_1)] & = & \frac{1}{n_1} \| \rvgamma^\star \mX \bar{\mH} \mDelta_\rvgamma \rvdelta^\star \|^2 + \frac{1}{n_1} \E[\|\rvgamma^\star \mX \bar{\mH} \rvvarepsilon \|^2] + \sigma^2 \nonumber \\
    & = & \frac{1}{n_1}  \| \rvgamma^\star \mX \bar{\mH} \mDelta_\rvgamma \rvdelta^\star \|^2 + \frac{\sigma^2}{n_1} \text{tr}(\bar{\mQ}_a^{-1} \mQ_1)+ \sigma^2. \label{eq:risk_pi_AG}
\end{eqnarray}

\clearpage

\section{Experimental Details}
\label{sec:experimental_details}

We now report the main details of all our experiments. All our experiments, including the reimplementation of other noisy label methods, are built on the open-source \texttt{uncertainty\_baselines} codebase~\citep{nado2021uncertainty} and follow as much as possible the benchmarking practices of \citet{ortiz2023does}.

\subsection{Datasets}
\label{app:dataset}

We use the following PI datasets to evaluate the performance of Pi-DUAL and other methods:

\textbf{CIFAR-10H} \citep{cifarh} is a relabeled version for CIFAR-10 \citep{krizhevsky2009learning} test set with 10,000 images. However, as proposed by \citet{collier2022transfer} we use CIFAR-10H as a training set so we use the standard CIFAR-10 training set as our test set. Following \citet{ortiz2023does}, we use the noisiest version of CIFAR-10H (denoted as ``worst") in our experiments. It has a noise rate (defined as the percentage of the labels that disagree with the original CIFAR-10 dataset) of approximately 64.6\%. The PI of CIFAR-10H consists of annotator IDs, annotator experiences and the time taken for the annotations. 

\textbf{CIFAR-10N and CIFAR-100N} \citep{cifarn} are relabeled versions of CIFAR-10 and CIFAR-100 with noisy human annotations. In our experiments, we use the noisiest version of these two datasets, known as CIFAR-10N (worst) and CIFAR-100N (fine), which both have a 40.2\% noise rate. The PI on these datasets consist on annotator IDs and annotator experience. It is worth noting that compared to CIFAR-10H, and as reported by \citet{ortiz2023does}, the PI on these two datasets is of a much lower quality. In general, it is much less predictive of the presence of a label mistake on a specific sample, as the PI features are only provided as averages over batches of samples.

\textbf{ImageNet-PI} \citep{ortiz2023does} is a relabeled version of the ImageNet ILSVRC12 dataset \citep{deng2009imagenet}. In contrast to the human-relabeled datasets described above, the labels of ImageNet-PI are provided by 16 different deep neural networks pre-trained on the original ImageNet. The PI for this dataset contains the annotator confidence, the annotator ID, the number of parameters of the model and its accuracy. In our experiments, we use both the high-noise (83.8\% noise rate) and low-noise version (48.1\% noise rate) of ImageNet-PI. 

A summary of the features of these datasets is given in Tab.~\ref{table:dataset}.

\begin{table}[h]
\centering
\caption{Summary of the main features of each of the datasets used in our experiments.}
\adjustbox{max width=\textwidth}{%
\begin{tabular}{cccccc}
    \toprule
                 & \begin{tabular}[c]{@{}c@{}}CIFAR-10H \\ {(worst)} \end{tabular} & \begin{tabular}[c]{@{}c@{}}CIFAR-10N\\ (worst)\end{tabular} & \begin{tabular}[c]{@{}c@{}}CIFAR-100N\\ (fine)\end{tabular}  & \begin{tabular}[c]{@{}c@{}}ImageNet-PI\\ (low-noise)\end{tabular} & \begin{tabular}[c]{@{}c@{}}ImageNet-PI\\ (high-noise)\end{tabular} \\
    \midrule
    Training set size  & 10k    & 50k     & 50k      & 1.28M                                                           & 1.28M                                                            \\
    PI quality   & High       & Low        & Low      & High                                                            & High                                                             \\
    Annotators    & Humans     & Humans      & Humans     & Models                                                           & Models                                                            \\
    Noise rate   & 64.6\%      & 40.2\%       & 40.2\%      & 48.1\%                                                            & 83.8\%                                                             \\
    \bottomrule
\end{tabular}
}
\label{table:dataset}
\end{table}

\subsection{Baselines} \label{app:baselines}

In our experiments, we compare the performance of Pi-DUAL on different tasks against several baselines selected to provide a fair comparison and good coverage of different methods in the literature. Specifically, we restrict ourselves to methods that only require training a single model on a single stage. We discard, therefore, methods that need multiple stages of training, such as Forward-T~\citep{patrini2017making} or Distillation PI~\citep{lopez2015unifying}; or multiple models, such as co-teaching~\citep{han2018co} and DivideMix~\citep{li2020dividemix}, as these are more computationally demanding, harder to tune, and in general harder to scale to the large-scale settings we are interested in. Also, most of these methods have been compared against more recent strategies like SOP~\citep{liu2022robust} or ELR~\citep{liu2020early}, and shown to perform worse than these baselines.

A short description of each of the baselines we compare to is provided below:

\textbf{Cross-entropy} Conventional training strategy consisting in the direct minimization of the cross-entropy loss between the model's predictions and the noisy labels.

\textbf{SOP}~\citep{liu2022robust} A noise-modeling method which models the label noise as an additive sparse signal. During training SOP uses the implicit bias of a custom overparameterized formulation to drive the learning of this sparse components. SOP needs $\mathcal{O}(n\times K)$ extra parameters over the cross-entropy baseline, where $n$ is the number of training samples, and $K$ the number of training classes.

\textbf{ELR}~\citep{liu2020early} This method adds an extra regularization term to the cross-entropy loss to bias the model's predictions towards their value in the early stages of training. To that end, it requires storing a moving average of the model predictions at each training iteration which adds $\mathcal{O}(n\times K)$ extra parameters over the cross-entropy baseline.

\textbf{HET}~\citep{het} Another noise-modeling method which models the uncertainty in the predictions as heteroscedastic per-sample Gaussian component in the logit space. The original version scales poorly with the number of classes, but the more recent HET-XL version~\citep{collier2023hetxl}, allows to scale this modeling approach to datasets with thousands of classes with only $\mathcal{O}(1)$ extra parameters coming from the small network that parameterizes the covariance of the noise. The standard version of HET achieves similar performance to HET-XL on ImageNet and can be run efficiently on this dataset. In our experiments, we thus use HET instead of HET-XL as a baseline.

\textbf{TRAM}~\citep{collier2022transfer} A PI-method which uses two heads, one with access to PI and one without it, to learn $p(\tilde{y}|\vx,\va)$ and $p(\tilde{y}|\vx)$ respectively. However, the feature extraction network leading to these heads is only trained using the gradients coming from the PI head. During inference, only the no-PI head is used. TRAM only requires $\mathcal{O}(1)$ extra parameters for the additional PI head.

\textbf{TRAM++}~\citep{ortiz2023does} On top of TRAM, TRAM++ augments the PI features with random sample-identifier to encourage the model to use the PI as a learning shortcut to memorize the noisy labels. 

\textbf{AFM}~\citep{collier2022transfer} Another PI-method that during training learns to approximate $p(\tilde{y}|\vx,\va)$ and during inference uses approximate marginalization based on the independence assumption $p(\va|\vx)\approx p(\va)$ and Monte-Carlo sampling to marginalize over $\va$. AFM only requires $\mathcal{O}(1)$ extra parameters to accomodate for the PI in the last layers.

\subsection{Hyperparameter Tuning Strategy}

As mentioned in Sec.~\ref{sec:experiments}, to ensure a fair comparison of the different methods, we apply the same hyperparameter tuning strategy in all our experiments and for all methods. In particular, we use a noisy validation set taken from the training set to select the best hyperparameters of a grid search. On CIFAR-10H, we randomly select $4\%$ of the samples; on CIFAR-10N and CIFAR-100N, $2\%$; and on ImageNet-PI, $1\%$ of all the samples in the training set. In all the experiments presented in the main text we use early stopping to select the best epoch to evaluate each method. Early stopping is also performed over the noisy validation set, although the reported accuracies are given over the clean test set.

\subsection{Training Details for CIFAR}
\label{app:train_details_cifar}

\textbf{General settings} We use a WideResNet-10-28 architecture in our CIFAR experiments. We train all models for $90$ epochs, with the learning rate decaying multiplicatively by $0.2$ after $36$, $72$ and $96$ epochs. We use a batch size of $256$ in all experiments, and train the models with an SGD optimizer with $0.9$ Nesterov momentum. In our grid searches, we sweep over the initial learning rate \{$0.01$, $0.1$\} and weight decay strength \{$10^{-4}$, $10^{-3}$\}. We always we use random crops combined with random horizontal flips as data augmentation.

\textbf{Method-specific settings} For ELR \citep{liu2020early}, we additionally sweep over the temporal ensembling parameter $\beta$ of \{0.5, 0.7, 0.9\} and the regularization coefficient $\lambda$ of \{1, 3, 7\}. For SOP \citep{liu2022robust}, we sweep over the learning rate for $u_i$ of \{1, 10, 100\}, as well as the learning rate for $v_i$ of \{1, 10, 100, 1000\}. We refer to the original papers for ELR \citep{liu2020early} and SOP \citep{liu2022robust} for detailed illustration of the hyperparameters.

For TRAM and AFM, we set the PI tower width to 1024 following the settings in \citet{collier2022transfer}. For TRAM++, we tune the PI tower width over a range of \{512, 1024, 2048, 4096\}. Additionally, we tune the random PI length of \{8, 14, 28\}, and no-PI loss weight over \{0.1, 0.5\} for TRAM++. 

For HET, we tune the heteroscedastic temperature over a range of \{0.25, 0.5, 0.75, 1.0, 1.25, 1.5, 2.0, 3.0, 5.0\}. For CIFAR-10H and CIFAR-10N, the number of factors for the low-rank component of the heteroscedastic covariance matrix is set to 3, while we set it to 6 for CIFAR-100N experiments. 

For Pi-DUAL, we set the width of the noise network and gating network to 1024 in the CIFAR-10H experiments, and to 2048 for the experiments with CIFAR-10N and CIFAR-100N. We use a three-layer MLP with ReLU activations for the noise network. The gating network shares the first layer with the noise network followed by another two fully-connected layers with ReLU activations and a sigmoid activation at its output.  We additionally search the random PI length over \{4,8,12,16\}. We do not apply weight decay regularization on the gating network and noise network for experiments on CIFAR for Pi-DUAL.\looseness=-1

We highlight that, compared with the competing methods (except the cross-entropy baseline), Pi-DUAL requires the smallest budget for hyperparameter tuning as it requires to tune only one additional hyperparameter, i.e., the random PI length; while the other methods require tuning at least two additional method-specific hyperparameters.

\subsection{Training Details for ImageNet-PI}

\textbf{General settings.} For all experiments with ImageNet-PI, we use a ResNet-50 architecture and SGD optimizer with Nesterov momentum of 0.9. The models are trained for 90 epochs in total with a batch size of 2048, with the learning rate decaying multiplicatively by 0.1 after 30, 60 and 80 epochs. The initial learning rate is set to 0.1, and we search over \{$10^{-5}$, $10^{-4}$\} for the weight decay strength. Random crop and random horizontal flip are used for data augmentation.

\textbf{Method-specific settings.} For all PI-related baselines (TRAM, TRAM++, AFM), we set the PI tower width to 2048. We set the no-PI loss weight to 0.5 and set the random PI length of 30 for TRAM++. For HET, we set the number of factors for the low-rank component of the heteroscedastic covariance matrix to 15, and we set the heteroscedastic temperature to 3.0. 

For Pi-DUAL, we set the random PI length to 30. The weight decay regularization on the gating network and noise network is the same as the prediction network. The architecture of the noise and gating network is the same as the one of the CIFAR experiments with a width of 2048.

\section{Additional Results}

\subsection{Training Dynamics on CIFAR-10N and ImageNet-PI (low noise)}
\label{app:training_dynamics}

\begin{figure}[ht]
\centering
\includegraphics[height=10cm]{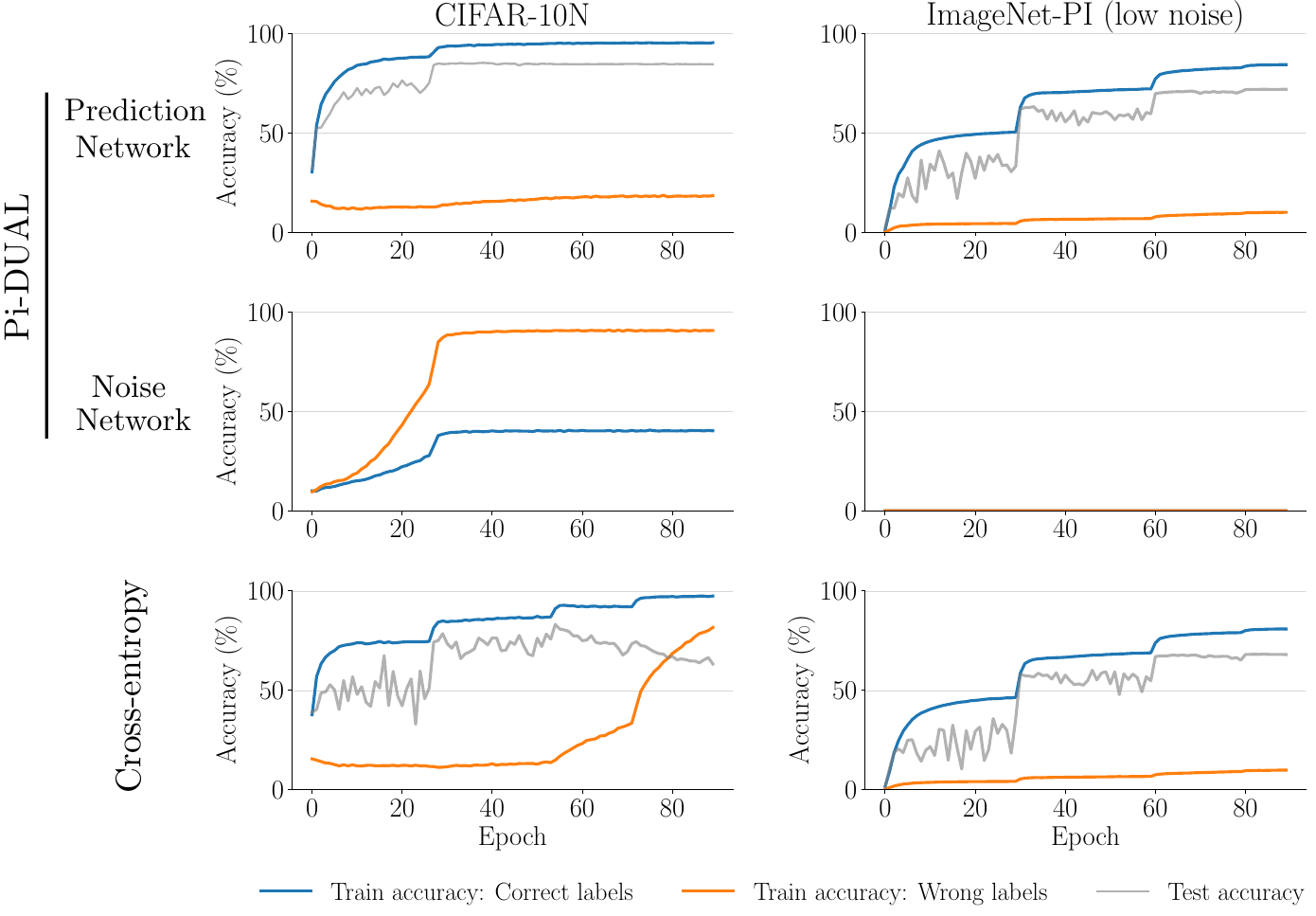}
\caption{Training accuracy dynamics on correct and mislabled samples, respectively for the prediction and noise sub-networks on CIFAR-10N (worst) and ImageNet-PI (low noise).}
\label{fig:training_dynamics_appendix} 
\end{figure}

Here we provide the training dynamics for CIFAR-10N and ImageNet-PI (low noise) in Fig. \ref{fig:training_dynamics_appendix} with the same findings as in Sec.~\ref{sec:dynamics}.

\subsection{Distribution of the Gating Network Predictions on CIFAR-10N and ImageNet-PI (low noise)}
\label{app:gate_distribution}

\begin{figure}[ht]
\centering
\includegraphics[width=0.65\textwidth]{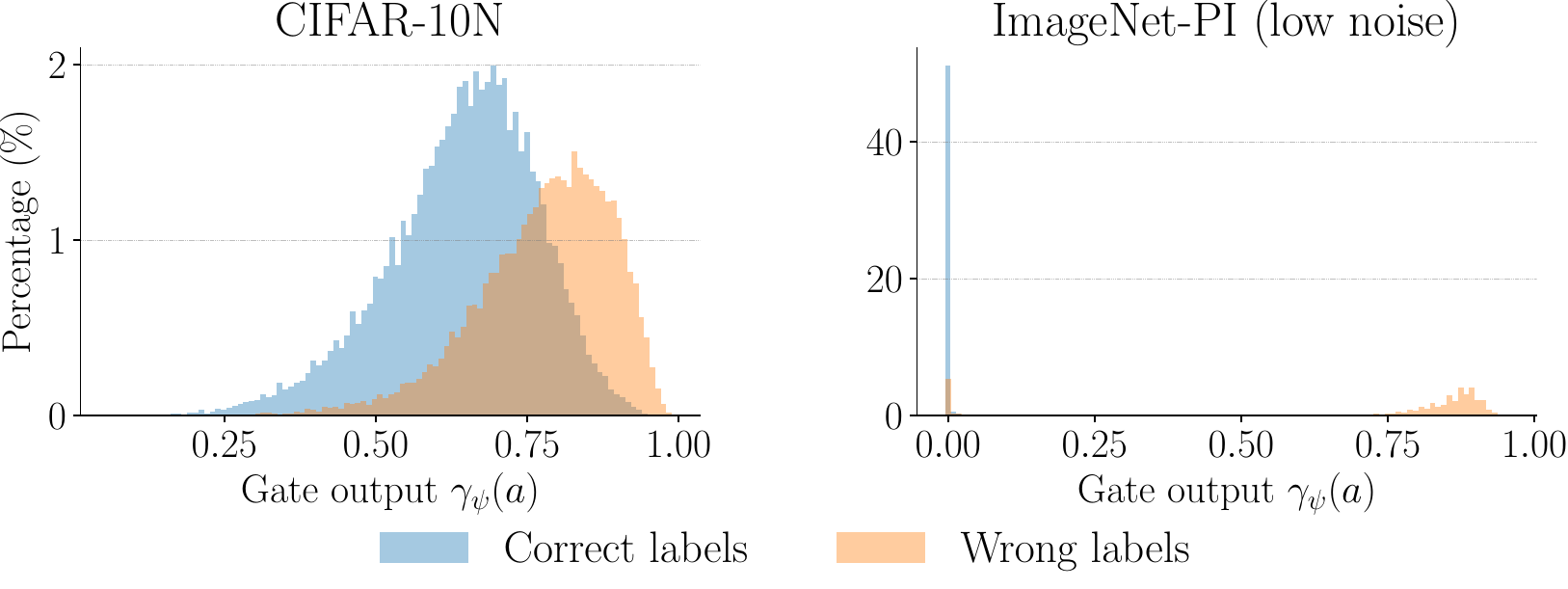}
\caption{Distribution of $\gamma_{\bm\psi}(\va)$, on correct training examples with correct and wrong labels for CIFAR-10N and ImageNet-PI (low noise).}
\label{fig:gate_distribution_appendix} 
\end{figure}

We show the distribution of the predictions of the gating network for CIFAR-10N and ImageNet-PI (low noise) in Fig. \ref{fig:gate_distribution_appendix} with the same findings as in Sec.~\ref{sec:gates}.

\subsection{Distribution of the Prediction Network Confidence on CIFAR-10N and ImageNet-PI (low noise)}
\label{app:confidence_distribution}

\begin{figure}[ht]
\centering
\includegraphics[width=\textwidth]{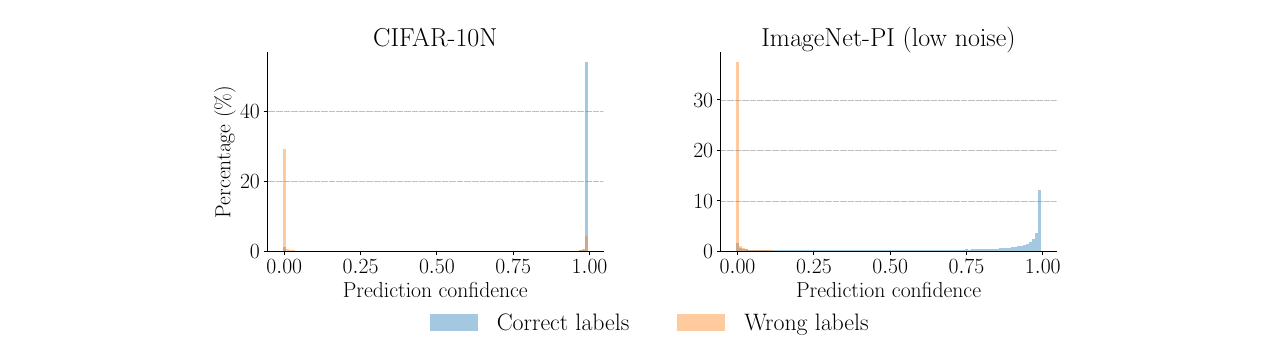}
\caption{Distribution for the prediction network's confidence on the observed noisy labels, separated by correctly and wrongly labeled samples.}
\label{fig:confidence_distribution_appendix} 
\end{figure}

We present the distribution of the prediction confidence of the prediction network on observed labels for CIFAR-10N and ImageNet-PI (low noise) in Fig. \ref{fig:confidence_distribution_appendix} complementing the findings of Sec.~\ref{sec:detection}. From the figure, we see that the confidence of the prediction network is clearly separated over samples with clean and wrong labels.

\subsection{Results without Early-stopping}
\label{app:no_early_stopping}

In the main text, as it is standard practice in the literature, we always provided results using early stopping. However, as mentioned before, early stopping is a key ingredient to achieve good performance by other methods, not Pi-DUAL. Indeed, Pi-DUAL still barely overfits to the incorrect labels using the prediction network, and thus it does not require early-stopping to achieve good results. To demonstrate this, Tab.  \ref{table:results_no_earlystop} reports the results of the same experiments as in Tab.~\ref{table:results}, but without using early stopping. 
From the table, we observe the performance of Pi-DUAL does not degrade in any of the datasets, while for other methods it suffers a heavily.

\begin{table}[ht]
\begin{center}
\caption{Test accuracy (without early stopping) on CIFAR-10H, CIFAR-10N, CIFAR-100N, and ImageNet-PI, comparing Pi-DUAL with previous methods (grouped by PI-based methods and No-PI methods). Mean and standard deviation are reported over 5 individual runs for CIFAR experiments and 3 runs for ImageNet experiments.}
\adjustbox{max width=\textwidth}{%
\begin{tabular}{@{}ccccccc@{}}
\toprule
& & & & & \\[-1em] %
\multicolumn{2}{c}{Methods}  & \begin{tabular}[c]{@{}c@{}}CIFAR-10H\\ (worst) \end{tabular} & \begin{tabular}[c]{@{}c@{}}CIFAR-10N\\ (worst)\end{tabular} & \begin{tabular}[c]{@{}c@{}}CIFAR-100N \\ (fine)\end{tabular} & \begin{tabular}[c]{@{}c@{}}ImageNet-PI\\ (low-noise)\end{tabular} & \begin{tabular}[c]{@{}c@{}}ImageNet-PI\\ (high-noise)\end{tabular} \\ \midrule
& & & & & \\[-1em] %
\multirow{3}{*}{\rotatebox[origin=c]{90}{No-PI}} 
                       &  Cross-entropy & $42.4_{\color{gray}\pm0.2}$ & $67.7_{\color{gray}\pm0.6}$ & $55.8_{\color{gray}\pm0.2}$ & $68.2_{\color{gray}\pm0.2}$ & $47.4_{\color{gray}\pm0.4}$ \\
                       &  ELR & $49.2_{\color{gray}\pm1.2}$ & $84.3_{\color{gray}\pm0.4}$ & $\textbf{63.8}_{\color{gray}\pm0.3}$ & - & - \\
                       &  SOP & $50.3_{\color{gray}\pm0.7}$ & $\textbf{86.0}_{\color{gray}\pm0.3}$ & $61.1_{\color{gray}\pm0.2}$ & - & - \\
                       \midrule
& & & & & \\[-1em] %
\multirow{3}{*}{\rotatebox[origin=c]{90}{PI}} & TRAM & $59.2_{\color{gray}\pm0.2}$ & $67.0_{\color{gray}\pm0.4}$ & $56.4_{\color{gray}\pm0.3}$ & $69.6_{\color{gray}\pm0.1}$ & $54.0_{\color{gray}\pm0.0}$ \\
                       &  TRAM++ & $64.7_{\color{gray}\pm0.6}$ & $82.3_{\color{gray}\pm0.1}$ & $60.6_{\color{gray}\pm0.2}$ & $69.5_{\color{gray}\pm0.0}$ & $54.1_{\color{gray}\pm0.1}$ \\
                       &  AFM & $61.2_{\color{gray}\pm0.7}$ & $69.8_{\color{gray}\pm0.5}$ & $58.9_{\color{gray}\pm0.3}$ & $70.3_{\color{gray}\pm0.0}$ & $55.3_{\color{gray}\pm0.2}$ \\ \midrule
& & & & & \\[-1em] %
\multicolumn{2}{c}{Pi-DUAL (Ours)} & $\textbf{73.8}_{\color{gray}\pm0.3}$ & $84.9_{\color{gray}\pm0.3}$ & $\textbf{64.2}_{\color{gray}\pm0.3}$ & $\textbf{71.7}_{\color{gray}\pm0.1}$ & $\textbf{62.3}_{\color{gray}\pm0.1}$ \\ \bottomrule

\end{tabular}%
}
\label{table:results_no_earlystop}
\end{center}
\end{table}

The same applies in the case of the noise detection results, where in Tab.~\ref{table:auc_number_no_earlystop} we see that Pi-DUAL can still detect the noisy labels equally well as in Tab.~\ref{table:auc_number} without the use of early stopping. The other methods on the other hand perform worse when applied to the last training epoch than to the early stopped one.

\begin{table}[t]
\caption{AUC of different noise detection methods without using early-stopping.}
\label{table:auc_number_no_earlystop}
\begin{center}

\adjustbox{max width=\textwidth}{%
\begin{tabular}{@{}cccccc@{}}
\toprule
& & & & & \\[-1em] %
  Methods & \begin{tabular}[c]{@{}c@{}}CIFAR-10H\\ (worst) \end{tabular} & \begin{tabular}[c]{@{}c@{}}CIFAR-10N\\ (worst)\end{tabular} & \begin{tabular}[c]{@{}c@{}}CIFAR-100N \\ (fine)\end{tabular} & \begin{tabular}[c]{@{}c@{}}ImageNet-PI\\ (low-noise)\end{tabular} & \begin{tabular}[c]{@{}c@{}}ImageNet-PI\\ (high-noise)\end{tabular} \\ \midrule
& & & & & \\[-1em] %
                       Cross-entropy & $0.558$ & $0.676$ & $0.666$ & $0.935$ & $0.941$ \\ 
                       ELR & $0.660$ & $0.839$ & $0.843$ & - & - \\ 
                       SOP & $0.743$ & $0.932$ & $0.793$ & - & - \\ 
                       TRAM++ & $0.887$ & $0.946$ & $0.890$ & $0.937$ & $0.959$ \\\midrule
& & & & & \\[-1em] %
                       Pi-DUAL (conf.) & $0.972$ & $\textbf{0.960}$ & $\textbf{0.910}$ & $\textbf{0.953}$ & $\textbf{0.987}$ \\ 
                       Pi-DUAL (gate) & $\textbf{0.983}$ & $0.815$ & $0.729$ & $\textbf{0.953}$ & $0.986$ \\ 
\bottomrule
\end{tabular}%
}
\end{center}
\end{table}

\subsection{Augmenting Pi-DUAL with State-of-the-art Regularization Techniques}
\label{app:pidual_plus}

In main text, we follow the common practice in literature and only compare Pi-DUAL with methods which do not incorporate techniques from semi-supervised learning, which greatly increases the computational cost and complexity. Here we propose an extension of Pi-DUAL, Pi-DUAL+, which boosts the performance of Pi-DUAL with additional regularization techniques.

Prior works in the literature have shown that the noisy-label training methods can be boosted with techniques from semi-supervised learning domain \citep{dividemix, liu2020early, liu2022robust}, at the cost of extra complexity costs and complexity. Pi-DUAL + adds to Pi-DUAL two regularization techniques, label smoothing and prediction consistency regularizer. 

\textbf{Label smoothing} Label smoothing is a regularization technique that was introduced to mitigate overconfidence during training by replacing hard labels with smoothed soft labels \citep{szegedy2016rethinking}. It has become a widely-used method to improve model generalization performance in classification tasks.

\textbf{Consistency regularizer} Prediction consistency regularizer is commonly used in both the semi-supervised learning \citep{berthelot2019mixmatch, sohn2020fixmatch, xie2020unsupervised} and learning with label noise literature \citep{cheng2020learning, liu2022robust}. It encourages the prediction consistency of the model across different input views. In Pi-DUAL+, we add a consistency regularizer $\mathcal{L}_C$ on the generalization term. Specifically, $\mathcal{L}_C$ is defined as the Kullback-Leibler divergence between softmax prediction from images with the default augmentation in Sec. \ref{app:train_details_cifar} and softmax predictions from the corresponding images augmented by Unsupervised Data Augmentation \citep{xie2020unsupervised}:

\begin{equation}
    \mathcal{L}_C = \frac{1}{N} \sum_{i=1}^{N}D_{\text{KL}} (\text{softmax}(f_{\bm\theta}(\bm x_i))\ \| \ \text{softmax}(f_{\bm\theta}(\text{UDA}(\bm x_i))))
\label{eq:loss_consistency}
\end{equation}

We use a hyper-parameter $\lambda_{C}$ to control the strength of the consistency regularizer.

For Pi-DUAL+, we sweep the label smoothing over \{0, 0.4\}, and $\lambda_{C}$ over \{0.5, 1\}. We train the Pi-DUAL+ for 300 epochs with a batch size of 128. The learning rate is set as 0.1 and decays with a cosine annealing schedule \citep{loshchilov2016sgdr}. Additionally, we sweep the random-pi length over \{4, 8\} and set the $l2$ regularization strength to $1e^{-4}$.

We compare Pi-DUAL+ against several semi-supervised learning pipeline methods, including Divide-Mix \cite{li2020dividemix}, CORES* \citep{cheng2020learning}, PES(semi) \citep{bai2021understanding}, ELR+ \citep{liu2020early} and SOP+ \citep{liu2022robust}. The results are compared in three datasets, CIFAR-10H, CIFAR-10N and CIFAR-100N, as shown in Tab.~\ref{table:results_semisupervised}. 

\begin{table}[ht]
\centering
\caption{Test accuracy on CIFAR-10H, CIFAR-10N and CIFAR-100N, comparing Pi-DUAL+ against state-of-the-art methods which combine noisy labels techniques with semi-supervised learning methods. The results of baseline methods on CIFAR-10N and CIFAR-100N are taken from \cite{wei2021learning, liu2022robust}}
\begin{tabular}{@{}cccc@{}}
\toprule
           & CIFAR-10H  & CIFAR-10N  & CIFAR-100N \\ \midrule
CE         & $51.10_{\color{gray}\pm2.20}$ & $80.60_{\color{gray}\pm0.20}$ & $60.40_{\color{gray}\pm0.50}$ \\
Divide-Mix & $71.68_{\color{gray}\pm0.27}$ & $92.56_{\color{gray}\pm0.42}$ & $\textbf{71.13}_{\color{gray}\pm0.48}$ \\
PES(semi)  & $71.16_{\color{gray}\pm1.78}$ & $92.68_{\color{gray}\pm0.22}$ & $70.36_{\color{gray}\pm0.33}$ \\
ELR+       & $54.46_{\color{gray}\pm0.50}$ & $91.09_{\color{gray}\pm1.60}$ & $66.72_{\color{gray}\pm0.07}$ \\
CORES*     & $57.80_{\color{gray}\pm0.57}$ & $91.66_{\color{gray}\pm0.09}$ & $55.72_{\color{gray}\pm0.42}$ \\
SOP+       & $66.02_{\color{gray}\pm0.06}$ & $\textbf{93.24}_{\color{gray}\pm0.21}$ & $67.81_{\color{gray}\pm0.23}$ \\ \midrule
Pi-DUAL+   & $\textbf{83.23}_{\color{gray}\pm0.26}$ & $\textbf{93.31}_{\color{gray}\pm0.21}$ & $67.99_{\color{gray}\pm0.08}$ \\ \bottomrule

\label{table:results_semisupervised}

\end{tabular}
\end{table}

From the table, we observe that, Pi-DUAL+ outperforms the other methods by a margin of over $11.0$ points on CIFAR-10H, while it performs on par with the state-of-the-art on the other two CIFAR-*N datasets. It demonstrates again the importance of good-quality PI features to maximize the performance of Pi-DUAL/Pi-DUAL+, and also demonstrates that Pi-DUAL can be effectively boosted by semi-supervised learning methods.

\subsection{Ablations over Model Structures}
\label{app:ablation_structure}

\subsubsection{Ablation for Prediction Network Backbone}

In all our CIFAR-level experiments (for both Pi-DUAL and other baselines methods), we used a WideResNet-10-28 as model backbone. Here we replace the WideResNet-10-28 with a ResNet-34 and present the performance comparison between Pi-DUAL and CE baseline on Tab.~\ref{table:ablation_backbone}.

\begin{table}[h]
\centering
\caption{Accuracy comparison between Pi-DUAL and CE on three datasets, using two different model backbones: WideResNet-10-28 and ResNet34.}
\label{table:ablation_backbone}
\begin{tabular}{@{}ccccc@{}}
\toprule
     & \multicolumn{2}{l}{WideResNet-10-28} & \multicolumn{2}{c}{ResNet34} \\ \midrule
Dataset \textbackslash Method & CE                & Pi-DUAL          & CE            & Pi-DUAL      \\ \midrule
CIFAR-10H                     & $51.1_{\color{gray}\pm2.2}$          & $\textbf{71.3}_{\color{gray}\pm3.3}$         & $51.4_{\color{gray}\pm2.1}$      & $\textbf{69.3}_{\color{gray}\pm3.1}$     \\
CIFAR-10N                     & $80.6_{\color{gray}\pm0.2}$          & $\textbf{84.9}_{\color{gray}\pm0.4}$         & $80.7_{\color{gray}\pm1.0}$      & $\textbf{84.5}_{\color{gray}\pm0.4}$     \\
CIFAR-100N                    & $60.4_{\color{gray}\pm0.5}$          & $\textbf{64.2}_{\color{gray}\pm0.3}$         & $56.9_{\color{gray}\pm0.4}$      & $\textbf{62.2}_{\color{gray}\pm0.3}$     \\ \bottomrule
\end{tabular}
\end{table}

From the results, we observe that Pi-DUAL maintains its performance improvement over CE with a different model backbone, exceeding the performance of CE baseline by a notable margin in all three datasets.

\subsubsection{Ablations on Structure for Noise and Gating Networks}

\textbf{Ablations on the width} In the paper we set the width of the PI-related modules (for both noise network and gating network) by default to 1024 for CIFAR-10H, and 2048 for all other experiments, without fine-tuning. Here we provide the results for using different widths for the PI networks on three CIFAR datasets in Tab. \ref{table:ablation_width}. 

\begin{table}[h]
\caption{Accuracy of Pi-DUAL on three datasets, varying the width of noise and gating networks of Pi-DUAL.}
\centering
\label{table:ablation_width}
\begin{tabular}{@{}ccccc@{}}
\toprule
Dataset \textbackslash \ Model width & 512      & 1024     & 2048     & 4096     \\ \midrule
CIFAR-10H                    & $\textbf{72.8}_{\color{gray}\pm2.9}$ & $\textbf{71.3}_{\color{gray}\pm3.3}$ & $\textbf{71.4}_{\color{gray}\pm3.6}$ & $\textbf{71.2}_{\color{gray}\pm3.8}$ \\
CIFAR-10N                    & $83.8_{\color{gray}\pm0.4}$ & $83.6_{\color{gray}\pm0.7}$ & $\textbf{84.9}_{\color{gray}\pm0.4}$ & $\textbf{85.3}_{\color{gray}\pm0.2}$ \\
CIFAR-100N                   & $62.3_{\color{gray}\pm0.2}$ & $63.7_{\color{gray}\pm0.4}$ & $\textbf{64.2}_{\color{gray}\pm0.3}$ & $\textbf{64.4}_{\color{gray}\pm0.2}$ \\ \bottomrule
\end{tabular}
\end{table}

From the table we observe that the performance of Pi-DUAL benefits from larger network width for the PI-related modules. 

\textbf{Ablations on the depth} In the paper we set the depth of the PI-related modules by default to 3 by default for all experiments, without fine-tuning. Here we provide the results for using different depths for the PI networks on three CIFAR datasets in Tab. \ref{table:ablation_depth}. Note that the width of the PI-related modules are set to the default value when tuning the depth.

\begin{table}[h]
\caption{Accuracy of Pi-DUAL on three datasets, varying the depth of noise and gating networks of Pi-DUAL.}
\centering
\label{table:ablation_depth}
\begin{tabular}{@{}cccc@{}}
\toprule
Dataset \textbackslash \ Model depth & 2 & 3 & 4 \\ \midrule
CIFAR-10H                             & $\textbf{68.9}_{\color{gray}\pm2.0}$   & $\textbf{71.3}_{\color{gray}\pm3.3}$   & $\textbf{70.1}_{\color{gray}\pm3.5}$   \\
CIFAR-10N                             & $83.4_{\color{gray}\pm0.4}$   & $\textbf{84.9}_{\color{gray}\pm0.4}$   & $\textbf{85.2}_{\color{gray}\pm0.3}$   \\
CIFAR-100N                            & $59.4_{\color{gray}\pm1.0}$   & $\textbf{64.2}_{\color{gray}\pm0.3}$   & $\textbf{64.1}_{\color{gray}\pm0.2}$   \\ \bottomrule
\end{tabular}
\end{table}

From the table we observe that the depth of the PI-related modules have to be at least of 3 layers to maximize the performance of Pi-DUAL.

\subsubsection{Ablations on Modeling of Noise Signals}

For Pi-DUAL, we chose to the model the label noise signal as a function of PI features $\bm{a}$. We show here that this modeling effectively prevents the model from memorizing the noise signal with input image features $\bm{x}$.

We show in Tab.~\ref{tab:ablation_noise_modeling} that the performance of Pi-DUAL deteriorates significantly if the noise signal is modeled as a function of both PI features $\bm{a}$ and input image features $\bm{x}$. This is because in this unconstrained setup, the model would directly memorize the label noise using the input images. This result demonstrates again that our modeling of label noise effectively decouples the learning of clean labels and the overfitting to the wrong labels.

\begin{table}[h]
\caption{Ablation for modeling of the noise signal.}
\centering
\label{tab:ablation_noise_modeling}
\begin{tabular}{@{}cccc@{}}
\toprule
Method \textbackslash \ Dataset & {CIFAR-10H} & {CIFAR-10N} & {CIFAR-100N} \\ 
\midrule
Cross-entropy & $51.1_{\color{gray}\pm2.2}$ & $80.6_{\color{gray}\pm0.2}$ & $60.4_{\color{gray}\pm0.5}$ \\
Pi-DUAL (model label noise with $\bm{a}$) & $\textbf{71.3}_{\color{gray}\pm3.3}$ & $\textbf{84.9}_{\color{gray}\pm0.4}$ & $\textbf{64.2}_{\color{gray}\pm0.3}$ \\
Pi-DUAL (model label noise with both $\bm{x}$ and $\bm{a}$) & $57.8_{\color{gray}\pm1.9}$ & $82.5_{\color{gray}\pm0.2}$ & $43.6_{\color{gray}\pm14.4}$ \\ 
\bottomrule
\end{tabular}
\end{table}

\subsection{Training Pi-DUAL with Loss Function of TRAM}
\label{app:pidual_tram_loss}

While both TRAM and Pi-DUAL utilize PI features to combat label noise, Pi-DUAL uses a much simpler loss function than TRAM, which implements a weighted combination of two functions to train its two heads \citep{collier2022transfer}. To validate the importance of the loss design of Pi-DUAL, which prevents the prediction network from overfitting to label noise, here we present the results for Pi-DUAL if we train it with the loss function of TRAM in Tab. \ref{table:pidual_tram_loss}.

\begin{table}[h]
\caption{Ablation for training Pi-DUAL using the loss function of TRAM.}
\centering
\label{table:pidual_tram_loss}
\begin{tabular}{@{}cccc@{}}
\toprule
Dataset \textbackslash \ Method & CE & Pi-DUAL & \begin{tabular}[c]{@{}c@{}}Pi-DUAL\\ with TRAM loss\end{tabular} \\ \midrule
CIFAR-10H                              & $51.1_{\color{gray}\pm2.2}$    & $\textbf{71.3}_{\color{gray}\pm3.3}$         & $61.6_{\color{gray}\pm4.8}$                                                                  \\
CIFAR-10N                              & $80.6_{\color{gray}\pm0.2}$    & $\textbf{84.9}_{\color{gray}\pm0.4}$         & $81.6_{\color{gray}\pm0.9}$                                                                  \\
CIFAR-100N                             & $60.4_{\color{gray}\pm0.5}$    & $\textbf{64.2}_{\color{gray}\pm0.3}$         & $59.0_{\color{gray}\pm0.6}$                                                                  \\ \bottomrule
\end{tabular}
\end{table}

\subsection{Performance of Pi-DUAL with Corrupted PI Features}

We have emphasized in the main text the importance of the quality of PI features to the performance of Pi-DUAL. In this section, we perform an ablation study where we gradually corrupt the PI features of CIFAR-10H, and train Pi-DUAL with the dataset with corrupted PI features. 

For each experiment, we randomly corrupt the PI features of a percentage of samples in the train set, where the corrupted PI features will be replaced by randomly generated PI features. Specifically, we substitute the annotator ID by a new random integer, and we substitute all other continual PI features by a random Gaussian vector with the same mean and standard deviation as the distribution of those features in the training set. We vary gradually the percentage of corrupted samples and report the performance for Pi-DUAL trained correspondingly in Tab. \ref{table:pi_corruption}.

\begin{table}[h]
\centering
\caption{Performance of Pi-DUAL with different levels of corruption in the PI features for the train set.}
\label{table:pi_corruption}
\begin{tabular}{@{}cccccc@{}}
\toprule
PI Corruption percentage & No corrupt & 25\% corrupt & 50\% corrupt & 75\% corrupt & 100\% corrupt \\ \midrule
Accuracy               & $\textbf{71.3}_{\color{gray}\pm3.3}$            & $66.4_{\color{gray}\pm2.7}$              & $63.9_{\color{gray}\pm0.7}$              & $52.6_{\color{gray}\pm4.0}$     & $47.2_{\color{gray}\pm3.7}$      \\ \bottomrule
\end{tabular}
\end{table}

From the table, we observe that the accuracy of Pi-DUAL decreases as there are more noise in the PI features of the training set, which demonstrates again the importance for high-quality PI features to maximize the performance of Pi-DUAL.

\subsection{Importance of Individual PI Features}

We study here the importance of each individual PI feature, by performing an ablation experiment on three CIFAR datasets where we remove one of the PI features while training Pi-DUAL.

The results are presented in Tab.~\ref{table:pi_importance}. It suggest that annotator ID is generally an important PI feature for these datasets, which suggests that the quality of the annotation varies from the annotators.
\begin{table}[h]
\centering
\caption{Ablation for the importance of each PI feature.}
\label{table:pi_importance}
\begin{tabular}{@{}cccc@{}}
\toprule
Method \textbackslash \ Dataset & {CIFAR-10H} & {CIFAR-10N} & {CIFAR-100N} \\ 
\midrule
Cross-entropy & $51.1_{\color{gray}\pm2.2}$ & $80.6_{\color{gray}\pm0.2}$ & $60.4_{\color{gray}\pm0.5}$ \\
Pi-DUAL (with all PI) & $\textbf{71.3}_{\color{gray}\pm3.3}$ & $\textbf{84.9}_{\color{gray}\pm0.4}$ & $\textbf{64.2}_{\color{gray}\pm0.3}$ \\
(w/o annotator ID) & $55.9_{\color{gray}\pm1.0}$ & $84.0_{\color{gray}\pm0.2}$ & $62.9_{\color{gray}\pm0.3}$ \\
(w/o annotator times) & $\textbf{73.6}_{\color{gray}\pm0.2}$ & $\textbf{84.7}_{\color{gray}\pm0.2}$ & $\textbf{64.2}_{\color{gray}\pm0.1}$ \\
(w/o random PI) & $\textbf{71.6}_{\color{gray}\pm2.9}$ & $82.1_{\color{gray}\pm0.3}$ & $61.5_{\color{gray}\pm0.4}$ \\
(w/o trial index) & $\textbf{74.4}_{\color{gray}\pm0.1}$ & {NA} & {NA} \\ 
\bottomrule
\end{tabular}
\end{table}

\section{Computational Cost in Large-scale Datasets}
\label{app:computational_cost}

In this paper, we used a TPU V3 with 8 cores for experiments on ImageNet-PI, and A100 (40G) for experiments on CIFAR.

Here we provide a computational cost analysis for Pi-DUAL on ImageNet-PI, comparing it with other baseline methods, with respect to both the number of parameters and the training time in Tab. \ref{table:training_cost}. Note that in the table we do not have run time for SOP and ELR as these two methods are very hard to scale to ImageNet-PI, where they require over 1 billion parameters. 

\begin{table}[h]
\centering
\caption{Computational cost analysis on ImageNet-PI in terms of number of parameters and running time, comparing Pi-DUAL with baseline methods.}
\label{table:training_cost}
\begin{tabular}{@{}ccccccc@{}}
\toprule
                     & CE     & TRAM++ & Pi-DUAL & HET    & SOP              & ELR              \\ \midrule
Number of parameters & 26M    & 32M    & 36M     & 58M    & \textgreater{}1B & \textgreater{}1B \\
Run time per step    & 0.510s & 0.541s & 0.566s  & 0.575s & -                & -                \\ \bottomrule
\end{tabular}
\end{table}

From the table, we can see that Pi-DUAL is a scalable method with almost the same training time as the cross-entropy baseline. Importantly, its parameters do not scale with neither the number of classes nor the number of samples, making it scalable to very large datasets.

\end{document}